\documentclass[conference]{IEEEtran}
\IEEEoverridecommandlockouts

\usepackage[numbers,sort&compress]{natbib}
\usepackage{amsmath,amssymb,amsfonts}
\usepackage{algorithm}
\usepackage{algpseudocode}
\usepackage{graphicx}
\usepackage{textcomp}
\usepackage{multirow}
\usepackage{subcaption}
\usepackage[table]{xcolor}
\usepackage{authblk}
\usepackage{balance}
\usepackage{listings}
\usepackage{stfloats}
\usepackage{caption}
\usepackage{tabularx}
\usepackage{float}
\usepackage{booktabs}
\usepackage{makecell}
\usepackage[hidelinks]{hyperref}
\newcolumntype{Y}{>{\centering\arraybackslash}X}

\def\BibTeX{{\rm B\kern-.05em{\sc i\kern-.025em b}\kern-.08em
    T\kern-.1667em\lower.7ex\hbox{E}\kern-.125emX}}

\newcommand{\model}{VisiFold}
\begin{document}

\title{VisiFold: Long-Term Traffic Forecasting via Temporal Folding Graph and Node Visibility}

\author[1, 3]{Zhiwei Zhang}
\author[2]{Xinyi Du}
\author[1, 3]{Weihao Wang}
\author[1, 3]{Xuanchi Guo}
\author[1, 3]{Wenjuan Han\thanks{\textsuperscript{*}Corresponding author.}\textsuperscript{*}}
\affil[1]{Key Laboratory of Big Data \& Artificial Intelligence in Transportation (Beijing Jiaotong University),}
\affil[ ]{Ministry of Education}

\affil[2]{Beijing Normal University, Beijing, China}
\affil[3]{School of Computer Science and Technology, Beijing Jiaotong University, Beijing 100044, China}

\affil[ ]{\texttt{\{zhiweizhang, weihaow, xuanchiguo, wjhan\}@bjtu.edu.cn,}}
\affil[ ]{\texttt{\{xinyidu\}@mail.bnu.edu.cn}}

\maketitle

\begin{abstract}
Traffic forecasting is a cornerstone of intelligent transportation systems. While existing research has made significant progress in short-term prediction, long-term forecasting remains a largely uncharted and challenging frontier. Extending the prediction horizon intensifies two critical issues: escalating computational resource consumption and increasingly complex spatial-temporal dependencies. Current approaches, which rely on spatial-temporal graphs and process temporal and spatial dimensions separately, suffer from snapshot-stacking inflation and cross-step fragmentation. To overcome these limitations, we propose \textit{VisiFold}. Our framework introduces a novel temporal folding graph that consolidates a sequence of temporal snapshots into a single graph. Furthermore, we present a node visibility mechanism that incorporates node-level masking and subgraph sampling to overcome the computational bottleneck imposed by large node counts. Extensive experiments show that VisiFold not only drastically reduces resource consumption but also outperforms existing baselines in long-term forecasting tasks. Remarkably, even with a high mask ratio of 80\%, VisiFold maintains its performance advantage. By effectively breaking the resource constraints in both temporal and spatial dimensions, our work paves the way for more realistic long-term traffic forecasting. The code is available at~\url{https://github.com/PlanckChang/VisiFold}.

\end{abstract}


\section{Introduction}

Traffic forecasting is a pivotal spatial-temporal task and plays a key role in an intelligent transportation system. As urban transportation networks expand rapidly, traffic management and route planning demand forecasts that look further into the future. Most existing studies focus on a short-term horizon of less than 1 hour, including STGNNs~\cite{STGNNs} and Transformer-based methods~\cite{pdformer, staeformer, st-transformer}. While these methods deliver strong short-term results, applying them directly to long-term forecasting leads to degraded accuracy and prohibitive computational costs~\cite{sstban}.

Earlier work also explored long-term forecasting~\cite{longterm1, longterm2, longterm3, long4, long5, long6}. They merely strive to improve forecasting performance, overlooking computational overhead, until SSTBAN~\cite{sstban}, which devises a bottleneck attention mechanism to reduce resource costs and introduces a multi-task learning framework to improve the performance. 

\begin{figure}[ht]
    \centering
    \includegraphics[width=1.1\linewidth]{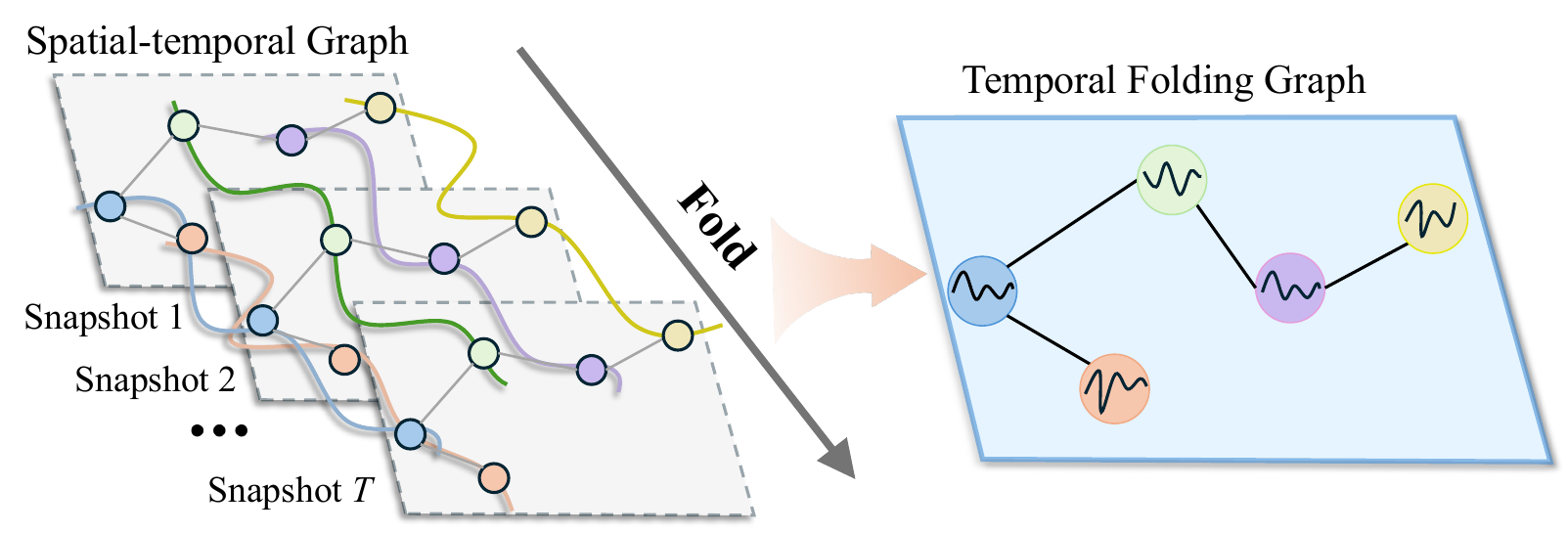}
    \caption{From spatial-temporal graph to temporal folding graph. A spatial-temporal graph leads to snapshot-stacking inflation and cross-step fragmentation, thereby constraining the expansion of the forecasting horizon. Temporal folding graph collapses all attributes over time steps into a single node, thereby compressing a sequence of snapshots into a single graph.}
    \label{fig:TFG}
\end{figure}

We rethink the conventional spatial-temporal graph modeling paradigm and find that it inherently degrades long-horizon performance and increases computational overhead. Concretely, it represents each temporal slice of the traffic network as a snapshot~\footnote{According to~\citeauthor{dynamicgraphSurvey}, spatial-temporal graphs are often framed as dynamic graphs composed of discrete static graphs. Bypassing minor terminological differences across fields, we use the term \textit{snapshot} to represent each of these temporal slices.} and uses a sequence of snapshots to capture temporal dynamics. This representation is intuitive and the cornerstone of progress in short-term forecasting. 

However, it naturally encourages spatial-temporal decoupling: information is aggregated within each snapshot by a spatial module and propagated across the temporal dimension by a temporal module.
Temporal dependencies are partitioned across separate snapshots and can be conveyed only through multiple intermediate representations, leading to cross-step fragmentation. Meanwhile, the resource overhead for a sequence of snapshots increases rapidly with the step size and the number of nodes in the traffic network, a phenomenon termed snapshot-stacking inflation.

To address the inherent limitations of the spatial-temporal graph, we introduce the \textbf{temporal folding graph} (TFG), as plotted in Fig.~\ref{fig:TFG}. Specifically, we embed all attributes (traffic signals in the physical world) for a node across multiple snapshots into a single, enriched attribute vector, akin to folding snapshots into a single graph. Thus, temporal dynamics are compressed within a node, and information interactions occur only once within a graph. TFG raises information density at the representation level, eliminating cross-step shuttling and propagation, and substantially reducing resource overhead.

While TFG escapes the forecast horizon restriction, the number of nodes in large road networks becomes the new bottleneck. We introduce \textbf{node visibility} to tackle this. Specific measures include node-level masking and subgraph sampling. The former randomly selects a subset of nodes and renders them invisible to the model. This shrinks the node size and discourages position-dependent bias
or overly tight interactions between nodes. The latter restricts each node’s receptive field to a smaller, randomly sampled neighborhood, thus increasing the parallelism.

In a nutshell, our contributions are fourfold:
\begin{enumerate}
    \item We revisit the root that limits forecasting horizons, the spatial-temporal graph representation, which enforces spatial-temporal decoupling modeling paradigm and leads to snapshot-stacking inflation and cross-step fragmentation.
    \item We propose the \textbf{temporal folding graph}, which compresses the node attributes along the temporal axis and encodes temporal dynamics within a single graph, thereby avoiding cross-step message passing and substantially reducing computational overhead while preserving long-term dependencies.
    \item To prevent the node size from becoming a bottleneck, we introduce \textbf{node visibility}, including node-level masking and subgraph sampling, enabling each node to interact only with a subset of peers; both accelerate training and save memory, and serve as an implicit regularizer that mitigates position dependence.
    \item Building on these ideas, we present \textbf{VisiFold} for long-term traffic forecasting that removes confinement in both temporal and spatial dimensions. Experiments show that \model~outperforms strong baselines on nine scenarios while accelerating training $>$$7\times$ and saving GPU memory $>$$4\times$.  
\end{enumerate}

\section{Related Work}
\subsection{Tokenization Technique}
TFG is an innovation in input tokenization and representation; accordingly, we compare it with works that focus on tokenization techniques.

The tokenization granularity determines the unit of information processed by models. To the best of our knowledge, there is no comparable technical approach to modify tokenization in traffic forecasting. Broadening to the general time series prediction, PatchTST~\cite{patchtst} and its successors~\cite{timexer, crossformer} segment the whole patch into sub-patches. TOTEM~\cite{totem} discretely quantizes the latent variable of time series data by a vector quantized technique~\cite{vqvae}. More generally, diverse tokenization techniques in NLP include character-level~\cite{character-level}, word-level~\cite{word2vec}, subword-level~\cite{bpe, bert}, sentence-level~\cite{sentencepiece}. ViT~\cite{vit} serves as the primary inspiration. Researchers processed pixel-level tokens directly in the age of CNN~\cite{unet, resnet} in CV. ViT, however, divides an image into block-level patches spanning multiple pixels and treats each patch as a token. Our TFG treats a node with a sequence of attributes merged from snapshots as a token.

\subsection{Spatial-Temporal Traffic Forecasting}
In early work, traffic prediction is treated as a multivariate time-series prediction task, and ARIMA~\cite{arima} and VAR~\cite{var} are intuitive methods. After exploring the spatial-temporal dependency~\cite{stgcn}, subsequent works adopt a separate paradigm to capture the dimensional characteristics. STGNNs~\cite{STGNNs} employ the GNN-like~\cite{gcn} spatial module for modeling graph topology. The typical temporal modules in STGNNs involve RNN-based~\cite{dcrnn, agcrn, stmetanet}, attention-based~\cite{gman, ts-at, pdformer, astgcn, astgnn, autost, geoman, stdn}, and CNN-based methods~\cite{stgcn, graphwarvenet, stsgcn, dmstgcn, stpgnn}. More recently, variants of Transformer~\cite{transformer} have begun to shine in the traffic prediction realm~\cite{sstban, pdformer, staeformer, st-transformer, hutformer}. Additionally, some studies~\cite{stgode, STGNCDE} focus on spatial information modeling. Besides, many efforts have been made to integrate differential equations with GNNs~\cite{dgcrn, ddSTGNN, dstagnn} and a normalization module~\cite{stnorm}. 

These existing models can be viewed as a technical route to improving model architecture, whereas our work focuses on the backbone's input representation.

\begin{figure*}
    \centering
    \includegraphics[width=1\linewidth]{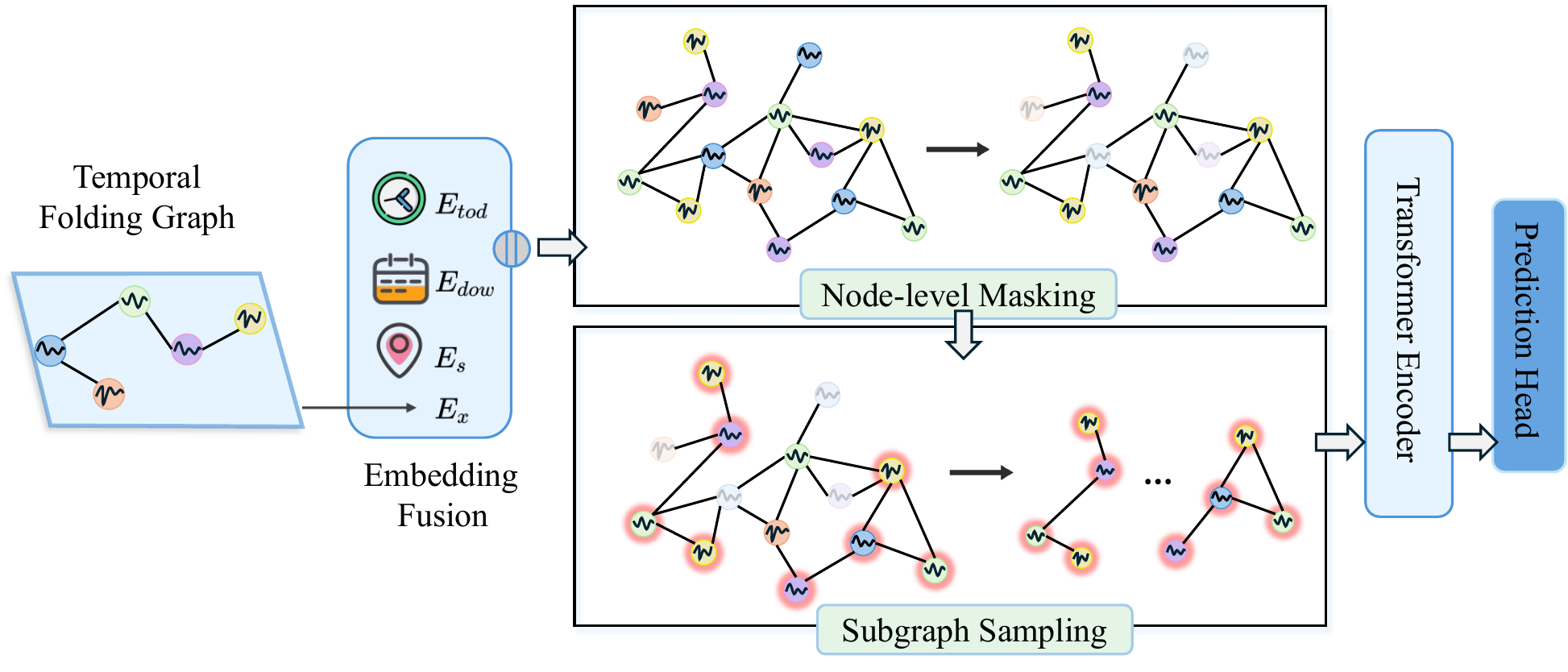}
    \caption{Overview of~\model.~\model~pipeline begins by building a temporal folding graph, from which token embeddings are derived via a linear transformation and fused with other embeddings. This is followed by node-level masking and subgraph sampling. The refined representations are then encoded by a Transformer encoder, and final predictions are generated by an MLP head.}
    \label{fig:overview}
\end{figure*}

\section{Method}
An overview of \model~is shown in Fig.~\ref{fig:overview}. In this section, we first describe the background of the traffic forecasting task. Then, we progressively break down our main contributions: the temporal folding graph and node visibility. 

\subsection{Problem Background}

\paragraph{Spatial-temporal Graph}  
A spatial-temporal graph represents the traffic network with time-varying traffic signals, including vehicle flow, speed, and road occupancy. The nodes $\mathcal{V}$ in the spatial-temporal graph denote the physical sensors deployed on the road, and edges $\mathcal{E}$ present connectivity.
Formally, the spatial–temporal graph consists of a sequence of snapshots and at timestamp $t$ the snapshot is $G_t:= (\mathcal{V}, \mathcal{E}, \mathbf{x}_t)$, where the $\mathbf{x}_t \in \mathbb{R}^{N\times C}$ denotes the attribute (traffic signal) of $N$ nodes.

\paragraph{Traffic Forecasting Task}
The traffic forecasting task forecasts how traffic signals in the traffic network change over time, using the observations from the past $T$ time steps to predict the future $T'$ time steps, formulated as:
\begin{align}
\hat{\mathbf{Y}}_{t+1:t+T'} &= f(\mathbf{X}_{t-T+1:t};\theta), \\
\mathbf{X}_{t-T+1:t}       &= [ \mathbf{x}_{t-T+1}, \mathbf{x}_{t-T+2}, \dots, \mathbf{x}_t ], \\ 
\hat{\mathbf{Y}}_{t+1:t+T'} &= [ \hat{\mathbf{x}}_{t+1}, \hat{\mathbf{x}}_{t+2}, \dots\hat{\mathbf{x}}_{t+T'} ],
\end{align}
where the $f$ is the target function with parameters $\theta$ to learn. 

\paragraph{Spatial-temporal Decoupling Paradigm}
The spatial-temporal graph implies spatial-temporal decoupling. 
Current models aggregate information within each snapshot and propagate it across time steps, leading to snapshot-stacking inflation. GPU memory and runtime grow rapidly with the horizon length. Meanwhile, spatial-temporal decoupling also induces cross-step fragmentation, undermining temporal dependencies and constraining long-term forecasting performance.

\subsection{Temporal Folding Graph}
We propose the temporal folding graph to address the snapshot-stacking inflation and cross-step fragmentation caused by the spatial-temporal decoupling paradigm accompanying spatial-temporal graphs.

Concretely, we embed all attributes across a sequence of snapshots into a single node, termed a TF-token. This representation avoids spatial-temporal decoupling; that is, information propagates on a single graph, increasing information density and enabling synchronized spatial-temporal modeling. 

In practice, for traffic forecasting, a scenario only needs to consider a single traffic signal type (i.e., C=1), so the channel dimension can be squeezed. Formally, given $\mathbf{X}_{t-T+1:t} \in \mathbb{R}^{N \times T \times C}$, we can define the $n$-th TF-token as:
\begin{align}
\mathbf{X}^n_{TF} & := \text{Squeeze}(\mathbf{X}_{t-T+1:t})[n] \in \mathbb{R}^{1\times T}
\end{align}
where $[\cdot]$ denotes the indexing operation. 
This implies that each node has $T$ attributes.

\subsection{Embedding Fusion}\label{sec:ternaryfunction}\label{sec:embeddingscheme}

 We then apply linear projection to obtain token embeddings $E_x \in \mathbb{R}^{N\times d}$, where $d$ is the hidden dimension.

Since the integration of spatial and temporal embedding has been proven to be a powerful technique for traffic forecasting~\cite{stgcn, graphwarvenet, gman, astgcn, astgnn, pdformer, staeformer}, we append additional information to the TF-token.
Spatial embeddings $E_s \in \mathbb{R}^{ N\times d}$ are initialized as a learnable matrix to identify each node. We determine temporal embeddings for the time-of-day and day-of-week cycles, whose periods are defined by the data frequency. These embeddings $E_{tod}$ and $E_{dow}$ are derived from the last timestamp $t$ of the input sequence and are shared across all nodes. The output of the embedding fusion is:
\begin{align}
E = E_x || E_s || E_{tod} || E_{dow} \in \mathbb{R}^{N \times 4d},
\end{align}
where $||$ is the concatenation operation.

\subsection{Node Visibility}

Notably, after representing the data as a temporal folding graph, the horizon constraint is lifted; however, the number of nodes still limits resource consumption. To address this, we introduce node visibility. Concretely, we apply node-level masking and subgraph sampling. We apply node visibility measures during training, but not during inference. The algorithm of node visibility is presented in Algorithm~\ref{alg:node_visibility}.

\paragraph{Node-level Masking}
Given a mask ratio $r$, we randomly select a proportion of tokens to mask. Specifically, following the design of MAE~\cite{mae}, we randomly remove nodes so they are unseen by the encoder. This directly reduces the number of input nodes, lowering resource consumption while preventing excessive mutual dependence among nodes. We also tried alternative masking strategies, such as the conventional approach of setting all or part of the selected nodes’ attributes to zero, but the results were unsatisfactory (see Section~\ref {sec:ablation}). Making entire nodes invisible is more economical and delivers better performance. 

\paragraph{Subgraph Sampling}
We randomly partition the remaining $(1-r)N$ nodes into subgraphs, each of size $s$. To ensure divisibility, we pad with $p$ nodes with all-zero attributes, resulting in $[(1-r)N + p]/{s}$ subgraphs. Partitioning into subgraphs further improves the model’s parallelism.

We gain the $\mathbf{Z}_0 \in \mathbb{R}^{\frac{(1-r)N+p}{s} \times s \times 4d}$ after the two operators.

In the general graph learning field, node-level masking and subgraph sampling are treated as structure-level augmentations to create different views for contrastive learning~\cite{graph_joco, graph-2, qiu2020gcc}. Unlike their approach, the primary purpose of our adopted node visibility is to circumvent the limitation on the number of nodes in large-scale graphs. Notably, we applied a very high masking ratio and observed that the performance still improved. We argue that in traffic networks, nearby nodes exhibit overly similar variation patterns, which allows the model to easily take shortcuts. By controlling the visibility level, we can force the model to learn more distant multi-hop dependencies, thereby increasing the difficulty of the task.

\begin{algorithm}
\caption{Node Visibility: Node-level Masking and Subgraph Sampling}
\label{alg:node_visibility}
\begin{algorithmic}[1]
\State \textbf{Input:} 
\State \quad Embedding matrix $E$,
\State \quad Original node set $\mathcal{V}$ with $|\mathcal{V}| = N$, \State \quad mask ratio $r$, 
\State \quad subgraph size $s$, 
\State \textbf{Output:}
\State \quad $\mathbf{Z}_0$
\\
\State \textbf{Step 1: Node-level Masking}
\State $\mathcal{V}_{\text{masked}} \gets \emptyset$
\State $M \gets \lfloor r \cdot N \rfloor$ \Comment{Number of nodes to mask}
\State Randomly select $M$ nodes from $\mathcal{V}$ to remove
\State $\mathcal{V}_{\text{remaining}} \gets \mathcal{V} \setminus \mathcal{V}_{\text{masked}}$ \Comment{Remaining $(1-r)N$ nodes}
\\
\State \textbf{Step 2: Subgraph Sampling}
\State $N_{\text{rem}} \gets |\mathcal{V}_{\text{remaining}}|$ \Comment{Number of remaining nodes}
\State $p \gets (s - (N_{\text{rem}} \mod s)) \mod s$ \Comment{Calculate padding needed}
\State $\mathcal{V}_{\text{padded}} \gets \mathcal{V}_{\text{remaining}} \cup \{\text{zero nodes}\}_{p}$ \Comment{Add $p$ zero-attributed nodes}
\State $K \gets (N_{\text{rem}} + p) / s$ \Comment{Number of subgraphs}
\State Partition $\mathcal{V}_{\text{padded}}$ into $K$ subgraphs $\{\mathcal{S}_1, \mathcal{S}_2, \dots, \mathcal{S}_K\}$
\State Each subgraph $\mathcal{S}_k$ contains exactly $s$ nodes
\\
\State \textbf{Step 3: Output}
\For{$k = 1$ to $K$}
\State Sample subgraph embedding $\mathbf{z}_k$ from $E$ with $\mathcal{S}_k$
\EndFor
\State $\mathbf{Z}_0 \gets \text{Concat}([\mathbf{z}_1, \mathbf{z}_2, \dots, \mathbf{z}_K])$  
\\ 
\State \textbf{Return:} 
\State \quad $\mathbf{Z}_0 \in \mathbb{R}^{\frac{(1-r)N+p}{s} \times s \times 4d}$ 
\end{algorithmic}
\end{algorithm}
\subsection{Network Architecture}
Based on the analysis above, we propose~\model. 
These four embeddings are concatenated and then used for node-level masking and subgraph sampling. Then, results are sent to a Transformer encoder, which consists of \(L\) layers of multi-head self-attention (MSA) and feed-forward network (FFN) blocks. Layer normalization (LN) is adopted before the block, and a skip connection is used afterward. Following the encoder, we attach an MLP as a prediction head, with GELU~\cite{gelu} as the non-linear activation function, to yield the output $\hat{\mathbf{Y}}$. The procedure of the encoder and prediction head can be formalized as:
\begin{align}
    \mathbf{Z^\prime}_\ell &= \operatorname{MSA}(\operatorname{LN}(\mathbf{Z}_{\ell-1})) + \mathbf{Z}_{\ell-1}, && \ell=1\ldots L \label{eq:msa_apply} \\
    \mathbf{Z}_\ell &= \operatorname{FFN}(\operatorname{LN}(\mathbf{Z^\prime}_{\ell})) + \mathbf{Z^\prime}_{\ell}, && \ell=1\ldots L  \label{eq:mlp_apply} \\
    \mathbf{\hat{Y}} &= \operatorname{MLP}(\mathbf{Z}_L) 
\end{align}
MSA can be further detailed as follows.
Given $\mathbf{Q}_i$, $\mathbf{K}_i$, and $\mathbf{V}_i$, gained through a linear mapping, serve as the inputs for $\text{head}_i$, the multi-head self-attention mechanism is
\begin{align}
 \text{MSA} & = \text{Concat}(\text{head}_1, \text{head}_2, ..., \text{head}_h)W^O \\
 \text{head}_i &= \mathrm{softmax}\left(\frac{\mathbf{Q}_i\mathbf{K}_i^T}{\sqrt{4d/h}}\right) \mathbf{V}_i
\end{align}
where $h$ is the number of heads, and $W^O$ denotes the output linear parameters.

Ultimately, the loss function is selected as Huber Loss~\cite{huber}, where $\delta$ is a positive value to trade off.
\begin{align}
    L(\mathbf{Y}, \hat{\mathbf{Y}}) = 
\begin{cases} 
\frac{1}{2}(\mathbf{Y} - \hat{\mathbf{Y}})^2 & \text{if } |\mathbf{Y} - \hat{\mathbf{Y}}| \leq \delta \\
\delta \cdot (|\mathbf{Y} - \hat{\mathbf{Y}}| - \frac{1}{2}\delta) & \text{otherwise}
\end{cases}
\end{align}

We provide a concise pseudocode of the whole process in Algorithm~\ref{alg:code}.

\begin{algorithm}
\caption{Algorithmic Procedure of~\model}
\label{alg:code}
\begin{algorithmic}[1]

\State \textbf{Input:} 
\State \quad Attribute matrix $\mathbf{X}_{t-T+1:t}$ 
\State \quad Timestamp $t$ ,
\State \quad Mask ratio $r$, 
\State \quad Subgraph size $s$,

\State \textbf{Output:} 
\State \quad $\hat{\mathbf{Y}}$

\\
\State \textbf{Step 1: TFG Construction and Embedding Fusion}
\State $\mathbf{X}^n_{TF} \gets \text{Squeeze}(\mathbf{X}_{t-T+1:t})[n]$ \Comment{Construct TFG}
\State $E_x \gets \text{Linear}(\mathbf{X_{TF}})$ \Comment{Linear transformation}
\State $E_s \gets \text{EmbeddingLookup}(\mathbb{R}^{N\times d}, n)$ \Comment{Lookup spatial embedding}
\State $E_{tod} \gets \text{Expand}(\text{EmbeddingLookup}(\mathbb{R}^{frequency \times d}, t))$ 
\State \quad \Comment{Lookup time-of-day embedding with the last  moment and share among all tokens}
\State $E_{dow} \gets \text{Expand}(\text{EmbeddingLookup}(\mathbb{R}^{7\times d}, t))$ 
\State \quad \Comment{Lookup day-of-week embedding with the last moment and share among all tokens}

\State $E \gets E_x \parallel E_s \parallel E_{tod} \parallel E_{dow}$ \Comment{Concatenate all embeddings}

\\
\State \textbf{Step 2: Node Visibility}
    \If{\textbf{When Training}}
        \State $\mathbf{Z}_0 \gets \Call{NodeVisibility}{E, \mathcal{V}, r, s}$
        \State \Return $\mathbf{Z}_0$
    \Else
        \State \Return $\mathbf{Z}_0 = E$ \Comment{At test time, return directly}
    \EndIf

\\
\State \textbf{Step 3: Transformer Encoder}
\For{$\ell = 1$ to $L$} \Comment{Apply $L$ layers of transformer}
    \State $\mathbf{Z^\prime}_\ell \gets \operatorname{MSA}(\operatorname{LN}(\mathbf{Z}_{\ell-1})) + \mathbf{Z}_{\ell-1}$ \Comment{Multi-head self-attention}
    \State $\mathbf{Z}_\ell \gets \operatorname{FFN}(\operatorname{LN}(\mathbf{Z^\prime}_\ell)) + \mathbf{Z^\prime}_\ell$ \Comment{Feed-forward network}
\EndFor

\\
\State \textbf{Step 4: Prediction}
\State $\hat{\mathbf{Y}} \gets \operatorname{MLP}(\mathbf{Z}_L)$ \Comment{Final prediction using an MLP}

\\
\State \textbf{Return:} 

\State \quad $\hat{\mathbf{Y}}$

\end{algorithmic}
\end{algorithm}

\subsection{Complexity Analysis}\label{sec:complexityAnalysis}
In a spatial-temporal graph, the previous models use a spatial module, commonly the graph model~\cite{gcn, transformer}, to capture spatial dependencies for every snapshot, and a temporal module, usually the sequential model~\cite{rnn, lstm, gru, transformer}, to aggregate temporal dynamics across snapshots, which leads to snapshot stacking inflation. For simplicity, let $g(T)$ denote the time and space complexity of the temporal module, and $h(N)$ represent the time and space complexity of the spatial module. Generally, the overall time and space consumption can be derived as $\mathcal{O}(N\cdot g(T) + T\cdot h(N))$~\cite{jin2023trafformer}.

In \model, the temporal folding graph eliminates the need for cross-step interaction and the temporal module, and reduces the token count from $
N\times T$ to $N$, thereby lowering the time and space complexity by an order of magnitude. Thus, the time and space complexity are both $\mathcal{O}(h(N))$.
After introducing node visibility, the number of tokens decreases further, so the complexity drops to $\mathcal{O}\!\left(\frac{(1-r)N + p}{s} \cdot h(s)\right)$, where the subgraph size $s$ is constant, $p$ denotes the number of padding tokens. Given the $h(s) = s^2$ since Transformer is employed as backbone, the ultimate complexity is $\mathcal{O}\!\left((1-r)Ns + ps\right)$. 
Node visibility offloads the quadratic cost and parallelizes subgraph computation.

\section{Experiment}
\begin{table}[tbh]
\centering
\begin{tabular}{@{}lcccc@{}}
\toprule
\textbf{Dataset} & \#\textbf{Time Steps} & \textbf{Nodes} & \textbf{Time Range} & \textbf{Frequency}        \\ \midrule

PEMS04  & 16992      & 307   & 2018/01 -- 2018/02      & 288           \\
PEMS08  & 17856      & 170   & 2016/07 -- 2016/08      & 288            \\           
SEATTLE & 8760       & 323   & 2015/01 -- 2015/12      & 24          \\ \bottomrule
\end{tabular}%
\caption{Summary of datasets.}
\label{tab:dataset}
\end{table}

\subsection{Setup}
\paragraph{Datasets.}
We evaluate the performance of \model~in three real-world datasets, PEMS04, PEMS08 and SEATTLE.
As following SSTBAN~\cite{sstban}, both \(T\) and \(T'\) are set to 24, 36, and 48. 
The PEMS series is collected from the Caltrans Performance Measurement System~\cite{datasetCollection} at a 288/day sampling frequency, with raw data including flow, speed, and occupancy; we select the flow feature. The SEATTLE dataset comprises speed data collected by inductive loop detectors deployed on freeways in the Seattle area~\cite{seattle}. TABLE \ref{tab:dataset} provides detailed dataset information. Consistent with previous work~\cite{sstban, stgcn, staeformer}, we split each dataset into training, validation, and testing sets with a 6:2:2 ratio. For preprocessing, the z-score method is applied to the raw data, and the mean and standard deviation are computed from the training set.

\paragraph{Baselines.}
We carefully chose the following 12 representative baselines. Statistical methods: \textbf{HA} and \textbf{VAR}~\cite{var}. STGNNs: \textbf{DCRNN}~\cite{dcrnn}, \textbf{GWNet}~\cite{graphwarvenet}, \textbf{AGCRN}~\cite{agcrn}, \textbf{DMSTGCN}~\cite{dmstgcn} and \textbf{STPGNN}~\cite{stpgnn}. Attention-based and Transformer-based methods: \textbf{GMAN}~\cite{gman}, \textbf{SSTBAN}~\cite{sstban}, \textbf{STAEformer}~\cite{staeformer} and \textbf{STDN}~\cite{stdn}.
MLP-based: \textbf{STID}~\cite{stid}.

  \begin{table*}[th]
  \centering
  \resizebox{.8\textwidth}{!}{
  \begin{tabular}{@{}cl|ccc|ccc|ccc@{}}
  \toprule
  \multirow{2}{*}{} &  \multirow{2}{*}{\textbf{Method}}  & \multicolumn{3}{c|}{\textbf{24 Time Steps}}    & \multicolumn{3}{c|}{\textbf{36 Time Steps}}    & \multicolumn{3}{c}{\textbf{48 Time Steps}}     \\  
                           &       & RMSE  & MAE   & MAPE      & RMSE  & MAE   & MAPE      & RMSE   & MAE   & MAPE      \\      \midrule
  \multirow{13}{*}{\rotatebox{90}{PEMS04}}  
                           & HA          & 81.57 & 56.47 & 45.49     & 106.58& 76.01 & 68.84     & 127.28 & 93.37 & 94.62     \\
                           & VAR         & 41.09 & 27.19 & 21.42     & 45.44 & 30.48 & 24.51     & 49.46  & 33.50  & 27.28     \\
                           & DCRNN       & 42.86 & 28.70 & 21.23     & 51.40 & 33.78 & 27.10     & 57.85  & 38.26 & 33.73     \\
                           & GWNet       & 35.52 & 22.79 & 16.04     & 38.17 & 24.71 & 17.67     & 40.60  & 26.42 & 18.99     \\
                           & GMAN        & 38.10 & 21.67 & 17.78     & 52.86 & 22.12 & 16.43     & 47.85  & 23.35 & 17.98     \\
                           & AGCRN       & 34.44 & 21.63 & 14.65     & 38.19 & 24.15 & 16.33     & 38.26  & 24.18 & 16.31     \\
                           & DMSTGCN     & 32.09 & 20.32 & 14.13     & 34.86 & 22.47 & 15.86     & 35.05  & 22.50 & 16.56     \\
                           & SSTBAN      & 32.82 & 20.17 & 14.43     & 34.15 & 20.82 & 14.83     & 35.51  & 21.66 & 15.90     \\
                           & STID        & \underline{31.57} & 19.48 & 13.28     & \underline{32.84} & 20.32 & 14.12     & \underline{33.95}  & \underline{21.09} & 14.59     \\
                           & STAEformer  & 31.71 & \underline{19.41} & \underline{12.68}     & \underline{32.84} & \underline{20.07} & \underline{12.94}    & 34.56  & 21.18 & \underline{13.78}     \\
                           & STPGNN & 47.03 & 31.13 & 22.61 & 52.78 & 34.61 & 24.05 & 59.69 & 40.23 & 30.29      \\

                           & STDN & 32.75  & 19.83 & 13.70 & 36.29 & 20.52		&13.72 & 35.31 & 21.28		& 14.02 \\
                           
                           \rowcolor{gray!20} \cellcolor{white}

                        & \textbf{\model}   & \textbf{31.36} & \textbf{19.07} & \textbf{12.44} & \textbf{32.52} & \textbf{19.78} & \textbf{12.92} & \textbf{33.23} & \textbf{20.27} & \textbf{13.35} \\
                            \midrule

  \multirow{13}{*}{\rotatebox{90}{PEMS08}}  
                           & HA          & 69.72 & 48.30  & 32.09     & 92.72 & 65.99 & 46.64     & 111.85 & 81.51 & 61.29     \\
                           & VAR         & 44.47 & 28.31 & 19.53     & 48.96 & 31.70  & 22.56     & 52.14  & 34.51 & 25.28     \\
                           & DCRNN       & 33.34 & 22.60 & 15.46     & 39.37 & 25.82 & 18.53     & 45.64  & 30.47 & 25.10     \\
                           & GWNet       & 29.47 & 19.07 & 12.25     & 33.54 & 21.76 & 13.68     & 34.20  & 22.60 & 14.16     \\
                           & GMAN        & 34.29 & 17.38 & 15.66     & 35.89 & 17.21 & 16.33     & 48.54  & 18.70 & 16.81     \\
                           & AGCRN       & 28.05 & 17.45 & 11.25     & 30.96 & 19.39 & 12.73     & 31.11  & 19.46 & 12.88     \\
                           & DMSTGCN     & 26.55 & 16.75 & 11.44     & 28.50 & 18.15 & 12.64     & 28.94  & 18.34 & 12.93     \\
                           & SSTBAN      & 26.32 & 15.97 & 12.29     & 28.30 & 16.84 & 12.20     & 28.82  & 16.94 & 12.47     \\
                           & STID        & \underline{25.67} & \underline{15.50} & \underline{10.41}     & 26.97 & 16.29 & 10.91     & 28.06  & 17.00 & 11.67     \\
                           & STAEformer  & 25.97 & 15.63 & 10.58     & \underline{26.84} & 16.06 & \underline{10.63}     & \underline{27.89}  & 16.90 & \underline{11.54}     \\ 
                           & STPGNN      & 38.98 & 25.59 & 16.29 & 40.20  & 26.15 & 17.22 & 42.64 & 28.01 & 20.66  \\
                           & STDN  & 26.60  & 15.54	& 11.99	& 27.69   & \underline{15.97} & 11.23 	 & 28.32 & \underline{16.50} 		& 11.97 \\
                           \rowcolor{gray!20} \cellcolor{white}
                           & \textbf{\model} & \textbf{25.31} & \textbf{14.73} & \textbf{9.84} & \textbf{26.51} & \textbf{15.48} & \textbf{10.38} & \textbf{27.49} & \textbf{16.35} & \textbf{10.98} \\
                         \midrule
 \multirow{13}{*}{\rotatebox{90}{SEATTLE}}  
                        & HA & 11.86 & 8.08 & 26.54 & 12.35 & 8.50 & 27.68 & 12.30 & 8.53 & 27.76 \\
                        & VAR & 9.33 & 6.22 & 18.58 & 9.57 & 6.29 & 19.54 & 9.87 & 6.45 & 20.49 \\
                        & DCRNN & 7.97 & 4.37 & 14.04 & 8.38 & 4.60 & 14.41 & 18.63 & 4.73 & 14.91 \\
                        & GWNet & 7.84 & 4.28 & 14.06 & 8.18 & 4.60 & 15.12 & 8.35 & 4.67 & 15.04 \\
                        & GMAN & 7.84 & 4.13 & 12.88 & 8.10 & 4.23 & 12.95 & 8.09 & 4.26 & 13.26 \\
                        & AGCRN & 7.83 & 4.27 & 13.53 & 8.31 & 4.66 & 14.76 & 8.60 & 4.82 & 15.62 \\
                        & DMSTGCN & 7.59 & 4.08 & 13.51 & 7.98 & 4.31 & 14.31 & 8.20 & 4.49 & 14.86 \\
                        & SSTBAN & 7.72 & 4.05 & 12.69 & 7.83 & 4.11 & \underline{12.44} & \underline{7.88} & \underline{4.12} & \underline{12.25} \\
                        & STID & 9.17 & 5.15 & 16.11 & 10.01 & 5.88 & 20.12 & 9.90 & 5.73 & 19.47 \\
                        & STAEformer & 7.59 & 3.88 & 12.53 & \underline{7.68} & \underline{4.00} & 12.78 & \underline{7.88} & 4.28 & 12.98 \\
                        & STPGNN & 7.85 & 4.25 & 14.14 & 8.12 & 4.52 & 14.92 & 8.48 & 4.72 & 16.16 \\
                        & STDN & \underline{7.40} & \underline{3.86} 	& \underline{11.99} & 7.73 & 4.07		& 12.65 & 7.98 & 4.17		& 12.59 \\
                        \rowcolor{gray!20} \cellcolor{white}
                        & \textbf{\model} & \textbf{7.22} & \textbf{3.82}   & \textbf{11.87} & \textbf{7.35} & \textbf{3.92}   & \textbf{12.15} & \textbf{7.48} & \textbf{3.99}     & \textbf{12.21} \\
                           \bottomrule
  \end{tabular}
  }
    \caption{Performance comparison
  in long-term scenarios. The values of the three metrics are the smaller, the better. The best results are indicated in \textbf{bold}, and the second-best results are \underline{underlined}. \model~outperforms the baselines in the long-term task.
  }
  \label{longtermcomparison}
\end{table*}

Below are brief descriptions of each baseline:
\begin{enumerate}
    \item \textbf{HA}: Historical Average, which predicts the future traffic flow by averaging the historical data.
    \item \textbf{VAR}~\cite{var}: Vector Auto-Regression, which is a classical multivariate time series forecasting method.
    \item \textbf{DCRNN}~\cite{dcrnn}: Diffusion Convolutional Recurrent Neural Network, predicting with diffusion convolutions and GRU.
    \item  \textbf{GWNet}~\cite{graphwarvenet}: Graph WaveNet, combining the dilated causal convolutions and graph convolutions to capture dynamics correlation.
    \item \textbf{AGCRN}~\cite{agcrn}: Adaptive Graph Convolutional Recurrent Network, which learns node-specific patterns and avoids predefined graphs.
    \item \textbf{DMSTGCN}~\cite{dmstgcn}: Dynamic and Multi-faceted spatial-temporal Graph Convolution Network, which captures the dependency with a spatial learning method and multi-faceted fusion module enhancement.
    \item \textbf{STPGNN}~\cite{stpgnn}: spatial-temporal Pivotal Graph Neural Networks, focusing on the pivotal nodes that show an important connection with other nodes to address the traffic dependencies. 
    \item  \textbf{GMAN}~\cite{gman}: Graph Multi-Attention Network, especially using spatial attention with randomly partitioned vertex groups.
    \item \textbf{SSTBAN}~\cite{sstban}: Self-Supervised spatial-temporal Bottleneck Attentive Network, utilizing a bottleneck attention scheme to reduce the computational cost.
    \item \textbf{STAEformer}~\cite{staeformer}: spatial-temporal Adaptive Embedding Transformer, proposing the spatial-temporal adaptive embedding that enhances the performance of vanilla Transformer in traffic prediction.
    \item \textbf{STDN}~\cite{stdn}: Spatiotemporal-aware Trend-Seasonality, designing a trend-seasonality decomposition module to disentangle the trend and cyclical component.
    \item \textbf{STID}~\cite{stid}: Spatial and Temporal IDentity information, which distinguishes the varying samples over spatial and temporal dimensions.
\end{enumerate}

\paragraph{Metrics.} 
We select three widely used metrics: root mean square error (RMSE), mean absolute error (MAE), and mean absolute percentage error (MAPE).

\paragraph{Implementation and Training Details.}
\model~utilizes the Adam optimizer~\cite{adam} with the default learning rate of 0.0001, which would decay with predefined milestones. The batch size is set to 16. The $\delta$ of Huber loss~\cite{huber} equals 1. \model~is implemented in PyTorch, and all experiments were conducted on one NVIDIA 3090.

\subsection{Long-term Forecasting Performance}\label{sec:performance}

 The performance results are presented in TABLE~\ref{longtermcomparison}. The best results are highlighted in bold, while the second-best ones are underlined. 
 As for the standard deviation, see the Fig.~\ref{fig:tfg_boxplots}. 
 Our model achieves the best performance across all prediction horizons.
 STID and STAEformer share the second-place ranking overall on PEMS04 and PEMS08, while SSTBAN and STAEformer do on SEATTLE. This performance gap demonstrates the varying capabilities of existing baselines across different datasets and further highlights the generalization and stability of our model. It is worth noting that STPGNN seems to underperform in long-term scenarios except for the SEATTLE 24-step case. This indicates that developing a specialized model designed for the more complex dependencies in long-term tasks is necessary, rather than simply extending short-term strong models to long-term tasks. These results suggest that \model~has the advantage of capturing long-term dependencies and generally achieves SOTA results on long-term tasks.

\begin{figure}[thb]
\hfill
    \centering

    \begin{subfigure}[b]{\columnwidth}
        \centering
        \includegraphics[width=.9\columnwidth]{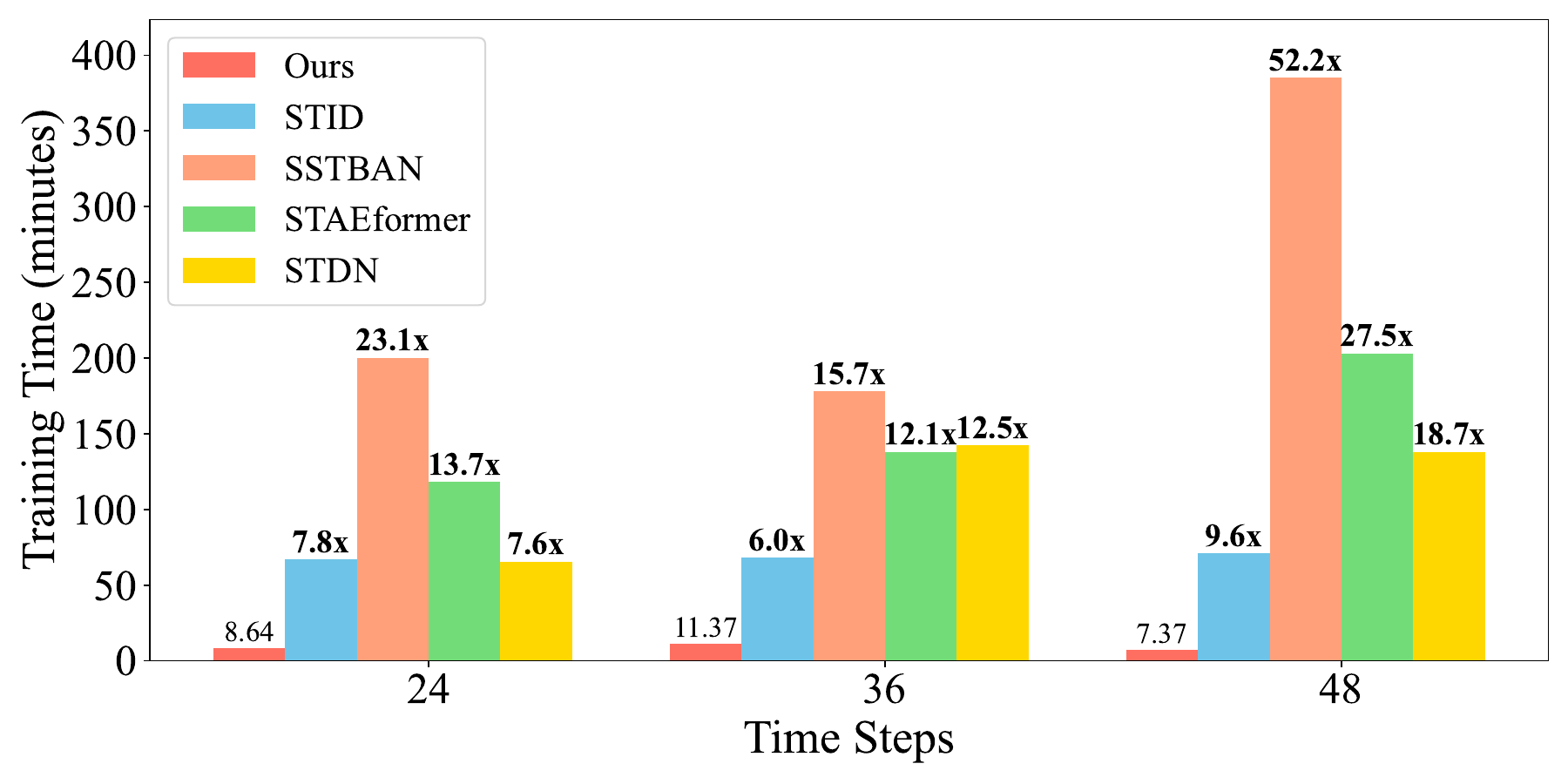}
        \caption{Training time.}
        \label{fig:trainingtimecomparison}
    \end{subfigure}

    \begin{subfigure}[b]{.9\columnwidth}
        \centering
        \includegraphics[width=\columnwidth]{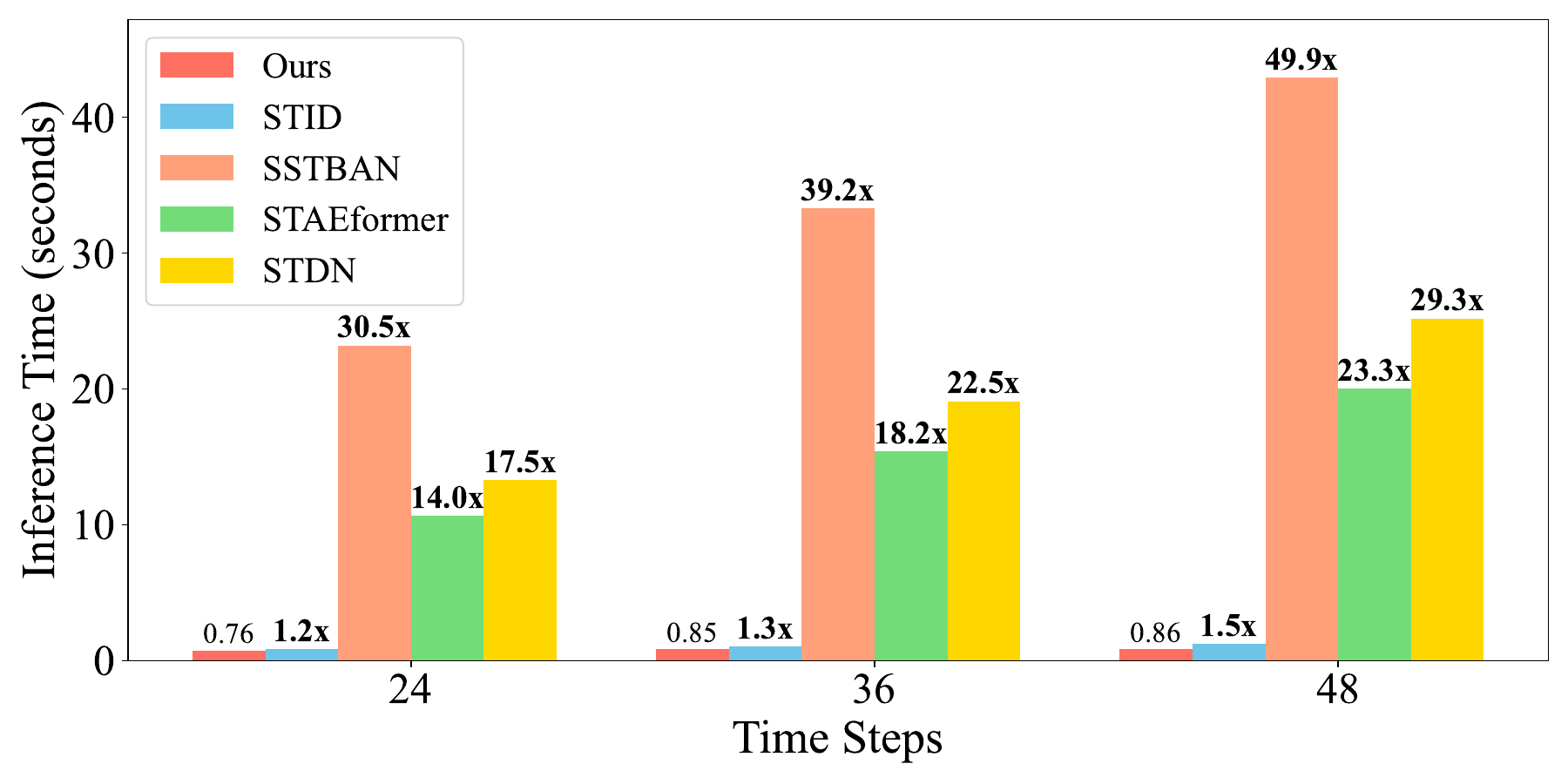}
        \caption{Inference time.}
        \label{fig:inferencetimecomparison}
    \end{subfigure}

    \begin{subfigure}[b]{.9\columnwidth}
        \centering
        \includegraphics[width=\columnwidth]{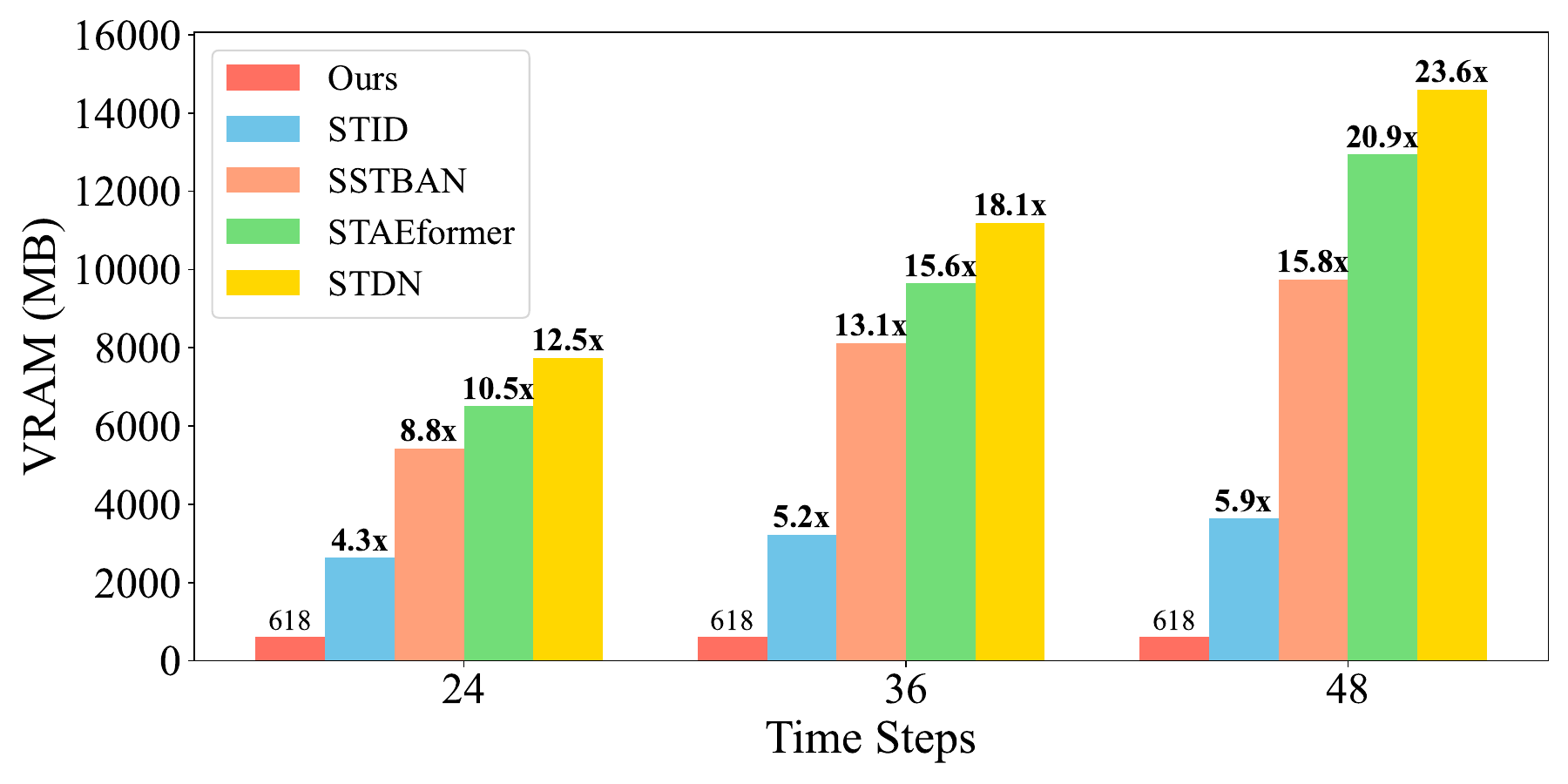}
        \caption{GPU memory.}
        \label{fig:memoryconsumption}
    \end{subfigure}
    
    \caption{Resource consumption comparison. The numbers above the bars for \texttt{Ours} are absolute values, while those above the baseline indicate multiples relative to \texttt{Ours}. \model~gains a significant superiority in resource overhead.}
    \label{fig:resourcecomparison}
\end{figure}

\subsection{Resource Consumption}\label{sec:resourceComparison}
We conducted a resource consumption evaluation in PEMS04 on the baselines whose performance is competitive in long-term tasks. The experimental results, shown in Fig.~\ref{fig:resourcecomparison}, detail changes in memory consumption, training time, and inference time as the time step increases from 24 to 48. Our model demonstrates significant superiority in terms of resource efficiency. 

Compared to the previous strongest model, STAEformer, \model~achieves a \textbf{17.8x} increase in training speed and a \textbf{18.5x} increase in inference speed while reducing memory consumption by \textbf{15.7x}, averaging across three different step-size settings. Compared to STID, the most resource-efficient baseline with strong performance, \model~improves average efficiency by \textbf{5.1x} in memory usage, \textbf{7.8x} in training time, and \textbf{1.3x} in inference time. As for other baselines, these improvements represent orders of magnitude in efficiency.

Since STID is based on an MLP approach, while our model uses a Transformer as its backbone, these advancements can be attributed to TFG and node visibility. Compared to SSTBAN, which is tailored for long-term tasks and utilizes bottleneck attention, our idea of modifying the representation without altering the backbone structure proves more efficient. Notably, our model's memory footprint increases marginally with increasing step length, enabling predictions over longer time horizons.

\subsection{Ablation Study and Analysis}\label{sec:ablation}

\paragraph{Ablation of the Additional Embeddings}

\begin{table}[ht]
\centering
\begin{tabular}{lccc}
\toprule
\textbf{Method} & \textbf{RMSE} & \textbf{MAE} & \textbf{MAPE} \\
\midrule
Full model & 31.36 & 19.07 & 12.44 \\
\midrule

w/o Spatial Embedding & +8.22 & +5.45 & +4.20 \\
w/o Dow Embedding     & +0.19 & +0.26 & +0.18 \\
w/o Tod embedding     & +0.35 & +0.29 & +0.27 \\
\bottomrule
\end{tabular}
\caption{Ablation of embedding components. The full model reports absolute errors in PEMS04 24-step scenario; ablated variants report the error increase (\,+\,) relative to the full model. While all embeddings are helpful, spatial embeddings provide the majority of the contribution.}
\label{tab:embedding_ablation}
\end{table}

The results of the ablation studies of embeddings on PEMS04 24-step are shown in TABLE~\ref{tab:embedding_ablation}. 
Removing $E_s$ results in the most significant degradation, indicating that spatial information is the dominant driver of accuracy. Temporal embeddings also help, but with more minor effects, where dropping $E_{tod}$ yields slightly larger increases than ablating $E_{dow}$. Overall, the results suggest all embeddings are helpful, and spatial embeddings carry most of the predictive power.

\paragraph{Ablation of Node Visibility}

\begin{figure*}[t]
  \centering
  \includegraphics[width=.9\linewidth]{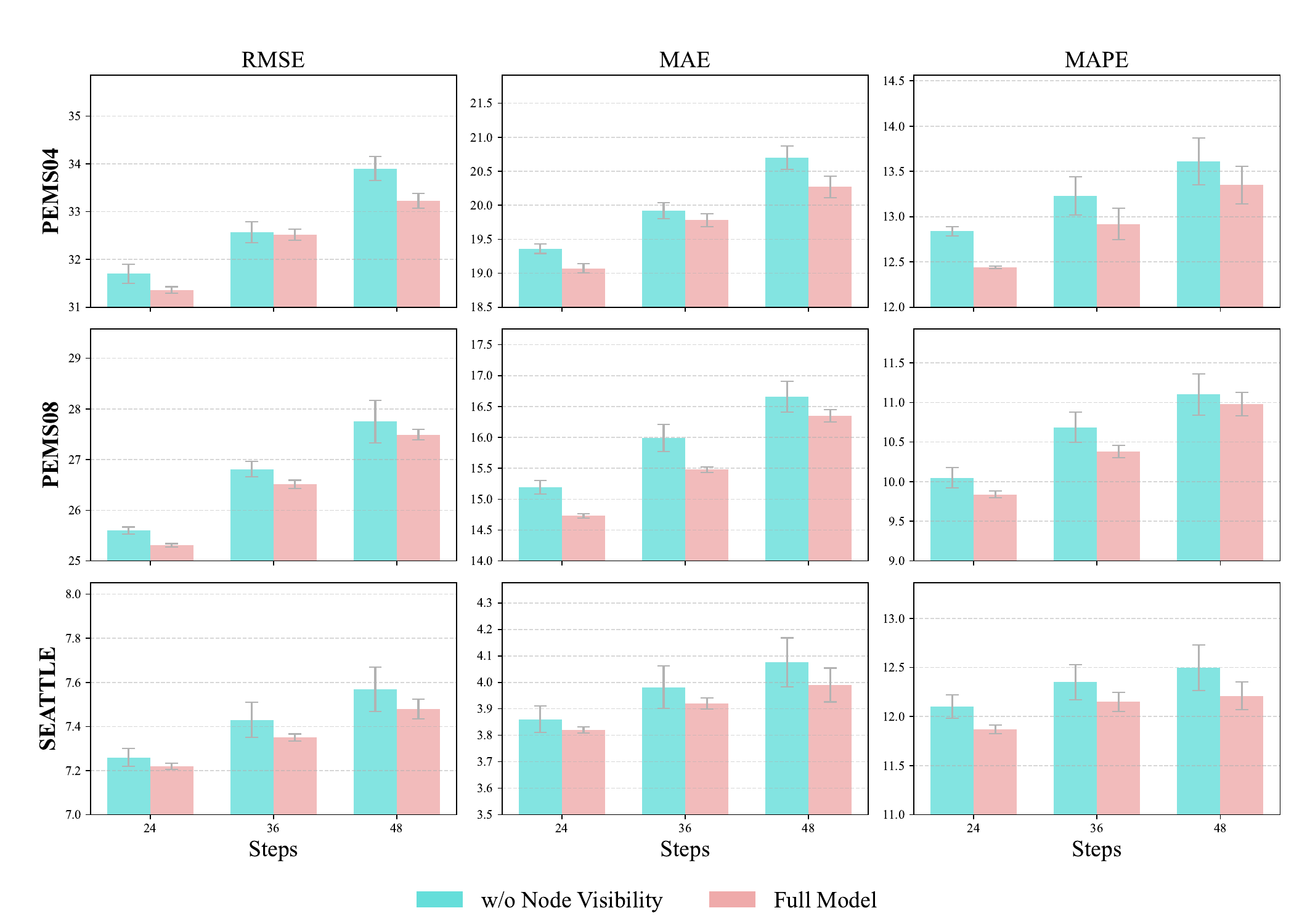}
  \caption{Performance and error bars of \texttt{Full Model} vs \texttt{w/o Node Visibility}. In the full version, which includes node-level masking and subgraph sampling, the average error decreases and the bar ranges narrow, indicating improved model stability and performance.}
  \label{fig:tfg_boxplots}
\end{figure*}

\begin{table}[ht]
    \centering
    \begin{tabular}{l|cccccc}
        \toprule
        & PEMS04 & PEMS08 & SEATTLE \\
        \midrule
        Only TFG                & 704   & 514   & 706   \\
        w/ Node-level Masking   & -10.23\% & -8.56\% & -6.23\% \\
        w/ Subgraph Sampling    & -1.70\%  & -2.33\% & -4.53\% \\
        Full Model              & -12.22\% & -10.86\% & -11.33\% \\
        \bottomrule
    \end{tabular}
    \caption{Memory overhead (MB) of the ablation of node visibility.  
    \texttt{Only TFG} uses absolute numbers, while the results with node visibility are reported as percentage reductions. }
    \label{tab:MBofNodeVisibilityablations}
\end{table}

We ablated node visibility and kept only TFG. We ran six experiments on each of three benchmarks and plotted boxplots for three metrics, as shown in Fig.~\ref{fig:tfg_boxplots}. Across all datasets and horizons, the full model shows lower medians and tighter interquartile ranges than the variant without node visibility, indicating it improved accuracy and greater stability. Errors increase with longer steps, as expected, with the deviation typically widening at 36 and 48 steps. SEATTLE exhibits lower absolute error, while PEMS04 and PEMS08 appear more challenging. Overall, the node visibility significantly contributes to the robustness and accuracy.

Regarding memory consumption, as shown in the TABLE~\ref{tab:MBofNodeVisibilityablations}, node visibility further reduces memory overhead.

\paragraph{Analysis of Mask Ratio}
\begin{figure*}[t]
  \centering
  \includegraphics[width=\linewidth]{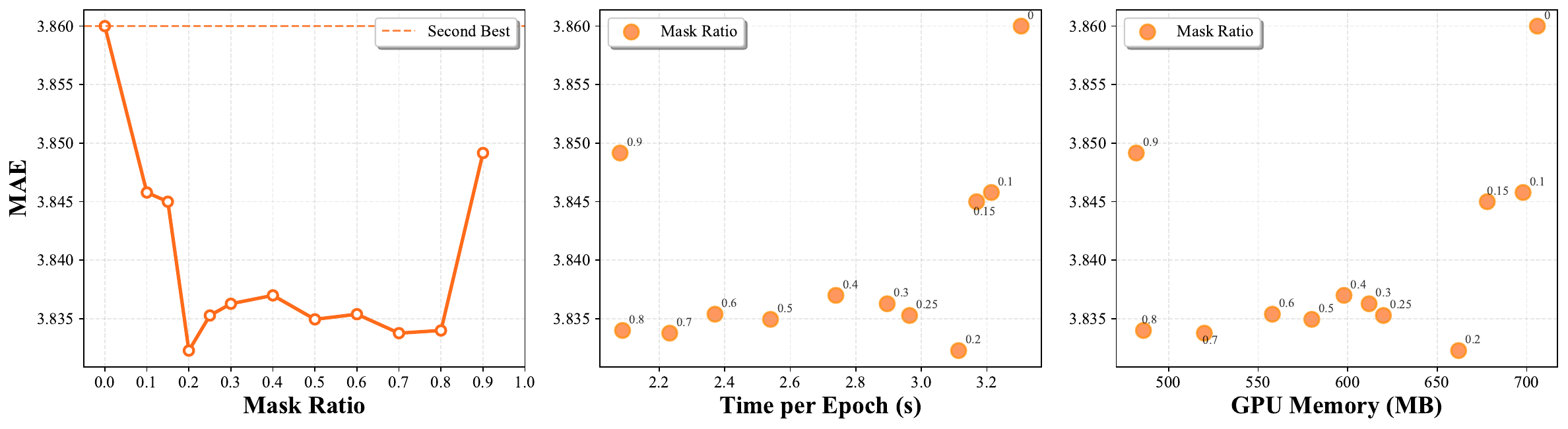}
  \caption{Analysis of mask ratio. We report the second-best performance from TABLE~\ref{longtermcomparison} as a reference. Even when 90\% of the nodes are masked, we still observe performance gains. During training, both per-epoch time and memory consumption decrease as the mask ratio increases.}
  \label{fig:maskratio}
\end{figure*}

We incorporate node-level masking into TFG rather than subgraph sampling to study the effect of the mask ratio in Fig.~\ref{fig:maskratio}. The experimental setting is 24-step forecasting on the SEATTLE dataset, with the mask ratio ranging from 0 to 0.9. The results show that node-level masking consistently improves model performance. Masking 20\% of the nodes achieves the best accuracy. When the mask ratio increases from 0.8 to 0.9, we observe an abrupt change in prediction accuracy. However, even with 90\% of the nodes masked, the model still improves performance compared to the baseline. This indicates substantial redundancy in traffic forecasting data.

From the perspective of resource consumption, both per-epoch time and memory usage decrease as the mask ratio increases. Meanwhile, we observe that the number of epochs required for convergence remains roughly constant. Therefore, on huge graphs with a large number of nodes, we can use a high mask ratio, such as 0.7 or 0.8, to achieve significant speedups and memory savings while maintaining prediction accuracy.

\paragraph{Masking Strategies Comparison}
We investigate different masking strategies, as shown in TABLE~\ref{tab:masking}. Using the SEATTLE 24-step scenario and disabling subgraph sampling, we tried three variants: (1) \texttt{AllZero}: set all node attributes to 0; (2) \texttt{PartialZero}: randomly set a subset of node attributes to 0; and (3) \texttt{RandomValue}: set node attributes to random values. We observe that all three potential approaches perform worse than node-level masking. We attribute this to a train–test gap: masking is not applied at test time, so the model may become misaligned and treat these visible perturbations as meaningful values. Considering that node-level masking also offers efficiency gains by making nodes invisible to the encoder, we choose it.

\begin{table}[h]
\centering
\begin{tabular}{lccc}
\toprule
\textbf{Method}       & \textbf{RMSE}  & \textbf{MAE}   & \textbf{MAPE}  \\
\midrule
Node-level   & 7.22  & 3.82  & 11.87 \\
\midrule
AllZero      & +0.08 & +0.05 & +0.05 \\
PartialZero  & +0.09 & +0.04 & +0.05 \\
RandomValue  & +0.15 & +0.07 & +0.14 \\
\bottomrule
\end{tabular}
\caption{Masking strategies comparison. Node-level masking surpasses other strategies.}
\label{tab:masking}
\end{table}

\begin{figure*}[th]
  \centering
  \includegraphics[width=\textwidth]{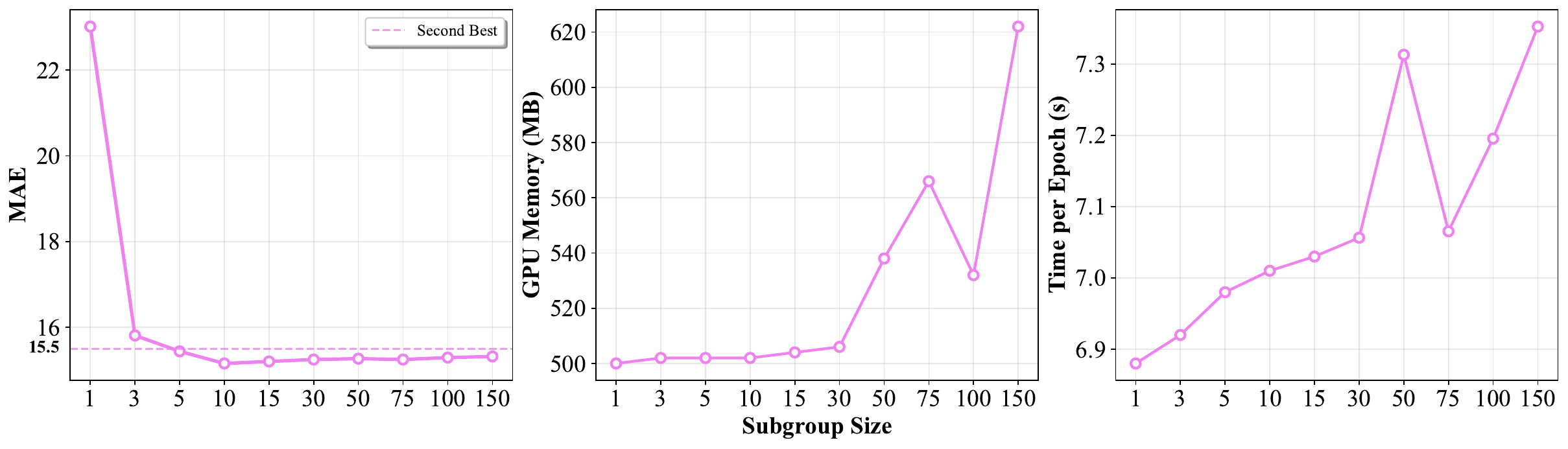} 
  \caption{Analysis of subgraph size. We report the second-best performance from TABLE~\ref{longtermcomparison} as a reference. When the subgraph is relatively small, resource consumption generally follows a linear pattern; however, as the subgraph grows large enough to affect padding values, fluctuations occur in the right part. Selecting subgraph size requires a balance between performance and efficiency. 
  }
  \label{fig:subgroup-metrics}
\end{figure*}

\paragraph{Analysis of Subgraph Size}
We investigated the impact of subgraph size. We set the mask ratio to 0 and ran experiments on PEMS08 with a 24-step horizon, as shown in Fig.~\ref{fig:subgroup-metrics}. The complexity analysis in Section~\ref{sec:complexityAnalysis} shows that the effect of subgraph size on computation speed and memory usage is linear, and the trend in the left part of the figure generally supports this conclusion. We also observed that when the subgraph size becomes comparable to the number of nodes (PEMS08 has 170 nodes), excessively large subgraphs lead to higher padding values, which in turn cause fluctuations in resource consumption at 50, 100, and 150, and it turns out that the subgraph size needs to be selected carefully. In terms of resource consumption, using smaller groups is preferable; however, this also reduces the number of nodes that can learn from each other, which is detrimental to overall generalization. Therefore, a tradeoff between performance and efficiency is required.

\subsection{Discussion}
\paragraph{Another Folding Strategy}

  \begin{table*}[th]
  \centering

  \resizebox{.75\textwidth}{!}{
  \begin{tabular}{@{}cc|ccc|ccc|ccc@{}}
  \toprule
  \multirow{2}{*}{} &  \multirow{2}{*}{\textbf{Method}}  & \multicolumn{3}{c|}{\textbf{24 Time Steps}}    & \multicolumn{3}{c|}{\textbf{36 Time Steps}}    & \multicolumn{3}{c}{\textbf{48 Time Steps}}     \\  
                           &       & RMSE  & MAE   & MAPE      & RMSE  & MAE   & MAPE      & RMSE   & MAE   & MAPE      \\      \midrule

  \multirow{2}{*}{PEMS04}  

                           & TFG & 31.70 & 19.36 & 12.84 & 32.57 & 19.92 & 13.23 & 33.90 & 20.70 & 13.61 \\
                           & SF      & 37.37 & 22.98 & 15.46      & 37.85 & 23.30 & 15.65     & 38.11 & 23.60 & 16.16 \\

                            \midrule
\multirow{2}{*}{PEMS08}  
                           & TFG & 25.60 & 15.19 & 10.05 & 26.81 & 15.99 & 10.69 & 27.75 & 16.66 & 11.10 \\

                           &SF       & 35.92 & 21.45 & 13.75        & 36.70 & 21.76 & 14.09  & 36.48 & 21.59 & 14.15  \\
                           \bottomrule
  \end{tabular}}
  \caption{Comparison of folding strategies. The \texttt{SF} counterpart folding along the spatial dimension inferior to the TFG.}
    \label{tab:TM&SM} \label{tab:adj}\label{tab:foldingComparision}
\end{table*}

Our TFG folds all attributes into the node along the temporal dimension. Naturally, there is a symmetric alternative: collapsing along the spatial dimension, termed SF. From the encoder’s computational perspective, the SF counterpart would only need to process a sequence of tokens equal to the time steps, which seems more efficient in traffic networks with hundreds of nodes.

We removed node visibility and focused purely on the folding strategy, running experiments on the PEMS04 and PEMS08 datasets. The results in the TABLE~\ref{tab:foldingComparision} show that TFG significantly outperforms SF. For SF, the main source of degradation is that we cannot seamlessly add spatial embeddings, because one SF token has multiple attributes from different nodes, whereas time-of-day and day-of-week embeddings can still be added. According to the embedding ablation study, spatial embedding is indispensable. Consequently, TFG provides a superior representation, with inter-node modeling for spatial dependencies and intra-node modeling for temporal correlations.

\paragraph{Potential Disruption to Topology}
Many prior studies place heavy emphasis on the topology/graph structure of traffic networks~\cite{stgcn, astgcn}, which employ the adjacency or learned topology to guide the information propagation. In contrast, node visibility explicitly disrupts structural dependence by randomly removing parts of the nodes and partitioning them into subgraphs. Yet, our empirical analysis shows that it not only reduces resource consumption but also improves performance.

\begin{figure}[t]
  \centering
    \includegraphics[width=.9\columnwidth]{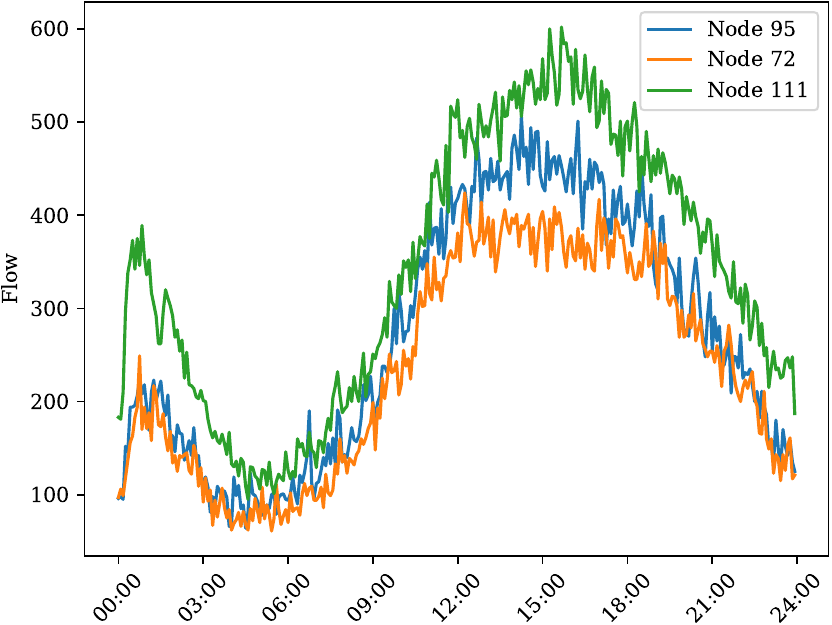}
    \caption{Illustration of traffic flow on the first day at Node 72, 95, and 111 in PEMS04. Node 72 and Node 95, being neighborhoods, exhibit similar flow patterns. Surprisingly, Node 111 also shows a similar trend, although it is not spatially adjacent to Node 72 and Node 95.}
    \label{fig:nodeflow}
\end{figure}

The inherent assumption of prior methods that emphasize topological relationships between nodes is that adjacent nodes have similar patterns, while distant nodes have different ones. In contrast, our visualization shows that similar patterns can emerge regardless of adjacency. We observed that many node pairs exhibit this characteristic, and one representative example is shown in Fig.~\ref{fig:nodeflow}, where both distant and nearby nodes can exhibit similar dynamics. This is aligned with the physical world: while traffic sensors are influenced by their neighbors, locations with similar functional roles in a city tend to share comparable traffic patterns. Meanwhile, recent studies~\cite{sstban, staeformer} have abandoned the use of predefined or learned topologies, instead entrusting neural networks with the direct learning of relational patterns between nodes. Furthermore, in the graph learning field, damaging the graph structure has also been proven to be an effective data augmentation method~\cite{graph_joco, graph-2}.

In summary, our extensive experiments demonstrate that node visibility can enhance model performance and reduce resource consumption. Node-level masking and subgraph sampling encourage the model to learn adjacency-insensitive representations, avoiding shortcut solutions and enhancing robustness and stability. In practice, road topology information is often difficult to obtain. Our work paves the way for dispensing with pre-specified topology, enabling more flexible learning.

\paragraph{Node-specific $E_{tod}$ and $E_{dow}$}
In VisiFold, $E_{tod}$ and $E_{dow}$, which indicate the time-of-day and day-of-week phase, are shared across nodes. This may overlook node-specific characteristics. Therefore, we also explore a exclusive variant for per node, and the results are reported in TABLE~\ref{tab:extraExp}. \texttt{Node-specific $E_{tod}$ and $E_{dow}$} adopt node-specific temporal embeddings, increasing the number of embedding parameters but yielding no performance gain.

\paragraph{Potential Loss of Global Dependencies}
Subgraph sampling aggregates features only locally, which may lead to the loss of global information. To analyze this, we introduced subgraph interaction scheme that a leader token within each subgraph to aggregate local representations, then performed cross-subgraph leader interactions to capture global dependencies. The updated leader tokens were subsequently propagated back to their corresponding subgraphs for an additional message-passing step. Specifically, \texttt{w/ Leader token (shared)} reuses the subgraph extraction network for leader-token communication, while \texttt{w/ Leader token (exclusive)} introduces an additional network dedicated to this interaction. While this mechanism is intuitive, our experiments in TABLE~\ref{tab:extraExp} showed that it did not improve performance while adding two extra forward passes. Therefore, despite seemingly enabling global dependency, this design provides limited practical benefit.

Beyond efficiency, we deem that allowing nodes to directly or indirectly access the entire graph may overly simplify the prediction task. As shown in Fig.~\ref{fig:nodeflow}, many nodes in traffic networks exhibit highly similar patterns, which increases the risk of overfitting to noise or spurious correlations. In contrast, our node visibility mechanism deliberately restricts the range of accessible features, challenging the traditional belief that learning a richer global topology is beneficial. Empirically, reducing visibility leads to more stable training and improved predictive performance, as shown in Fig.~\ref{fig:tfg_boxplots} and~\ref{fig:maskratio}.

\begin{table*}[h]
\centering
\begin{tabular}{l l|ccc|ccc|ccc}
\toprule
\multirow{2}{*}{\textbf{Dataset}} & \multirow{2}{*}{Setting} & \multicolumn{3}{c|}{\textbf{24 Time Steps}} & \multicolumn{3}{c|}{\textbf{36 Time Steps}} & \multicolumn{3}{c}{\textbf{48 Time Steps}} \\
 & & RMSE & MAE & MAPE & RMSE & MAE & MAPE & RMSE & MAE & MAPE \\
\midrule
\multirow{4}{*}{PEMS04} 
 & \textbf{VisiFold}    & \textbf{31.36} & \textbf{19.07} & \textbf{12.44} & \textbf{32.52} & \textbf{19.78} & \textbf{12.92} & \textbf{33.23} & \textbf{20.27} & \textbf{13.35} \\
& Node-specific $E_{tod}$ and $E_{dow}$ & 31.64 & 19.32 & 12.57 & 32.93 & 20.07 & 13.02 & 33.52 & 20.53 & 13.48 \\

& w/ Leader token (shared)  & 32.02 & 19.52 & 12.89 & 33.15 & 20.29 & 13.28 & 33.88 & 20.76 & 13.95 \\
& w/ Leader token (exclusive) & 32.03 & 19.62 & 13.09 & 32.92 & 20.16 & 13.92 & 35.01 & 21.22 & 13.80 \\
\midrule
\multirow{4}{*}{PEMS08} 
& \textbf{VisiFold}    & \textbf{25.31} & \textbf{14.73} & \textbf{9.84} & \textbf{26.51} & \textbf{15.48} & \textbf{10.38} & \textbf{27.49} & \textbf{16.35} & \textbf{10.98} \\
& Node-specific $E_{tod}$ and $E_{dow}$ & 26.06 & 15.42 & 10.08 & 27.27 & 16.29 & 10.67 & 27.77 & 16.58 & 11.02 \\

& w/ Leader token (shared)   & 25.58 & 15.45 & 10.25 & 26.61 & 16.00 & 10.68 & 27.56 & 16.41 & 11.05 \\
& w/ Leader token (exclusive) & 25.46 & 15.35 & 10.17 & 26.64 & 16.09 & 11.09 & 27.57 & 16.45 & 11.09 \\
\midrule
\multirow{4}{*}{SEATTLE} 
& \textbf{VisiFold}    & \textbf{7.22} & \textbf{3.82} & \textbf{11.87} & \textbf{7.35} & \textbf{3.92} & \textbf{12.15} & \textbf{7.48} & \textbf{3.99} & \textbf{12.21} \\
& Node-specific $E_{tod}$ and $E_{dow}$ & 7.38 & 3.99 & 11.93 & 7.43 & 3.94 & 12.23 & 7.57 & 4.06 & 12.41 \\

& w/ Leader token (shared)   &  7.47 &  3.98 & 12.90 &  7.63 &  4.11 & 13.28 &  7.63 &  4.15 & 13.18 \\
& w/ Leader token (exclusive) &  7.41 &  3.93 & 12.63 &  7.63 &  4.10 & 13.15 &  7.71 &  4.15 & 13.42 \\

\bottomrule
\end{tabular}
\caption{Analysis of node-specific temporal embeddings and subgraph interaction scheme.}
\label{tab:extraExp}
\end{table*}

\subsection{Visualization Analysis}
\begin{figure*}[]
    \centering
    \begin{subfigure}[b]{.33\textwidth}
        \includegraphics[width=\textwidth]{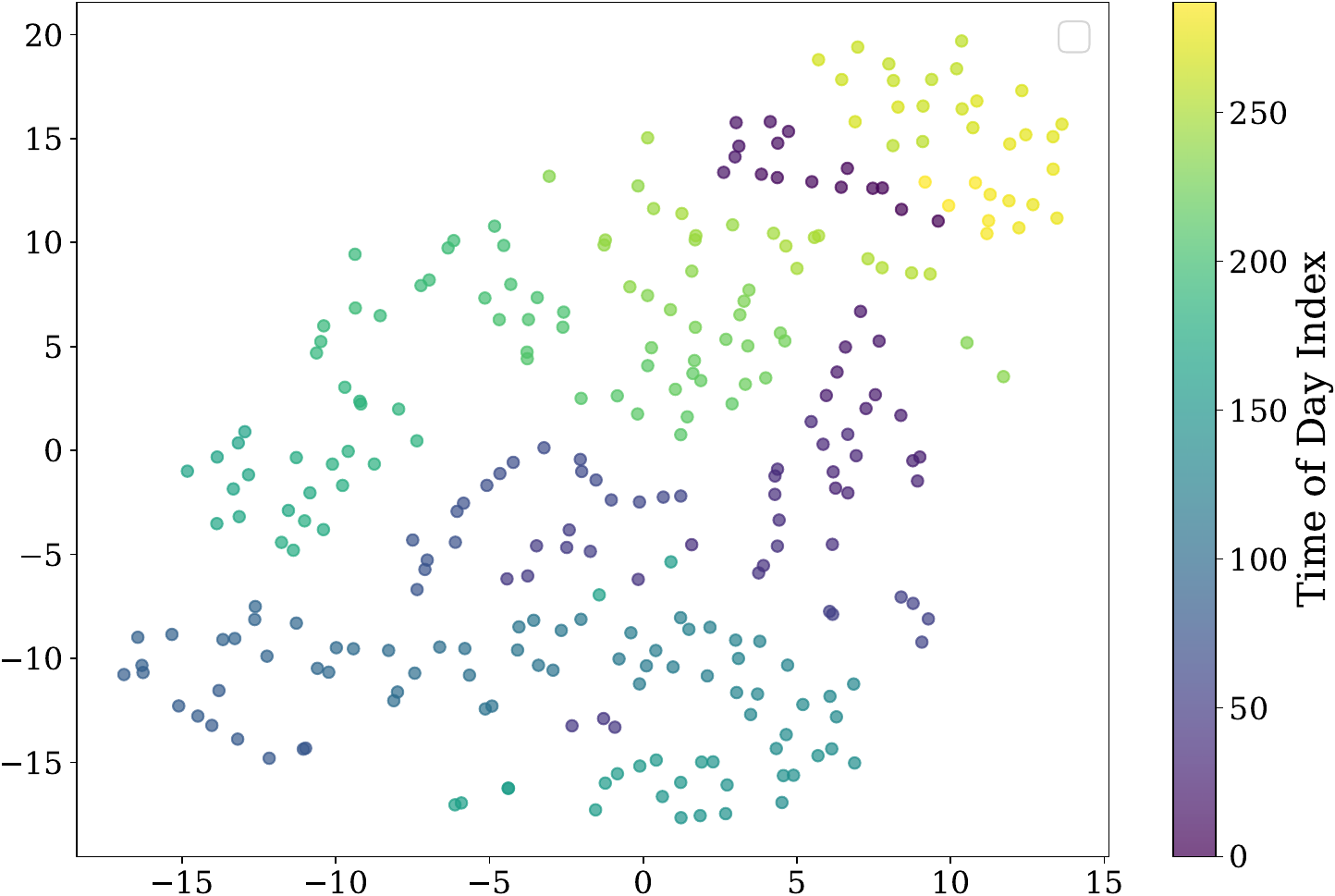}
        \caption{$E_{tod}$.}
        \label{fig:tod}
    \end{subfigure}
    \begin{subfigure}[b]{.30\textwidth}
        \includegraphics[width=\textwidth]{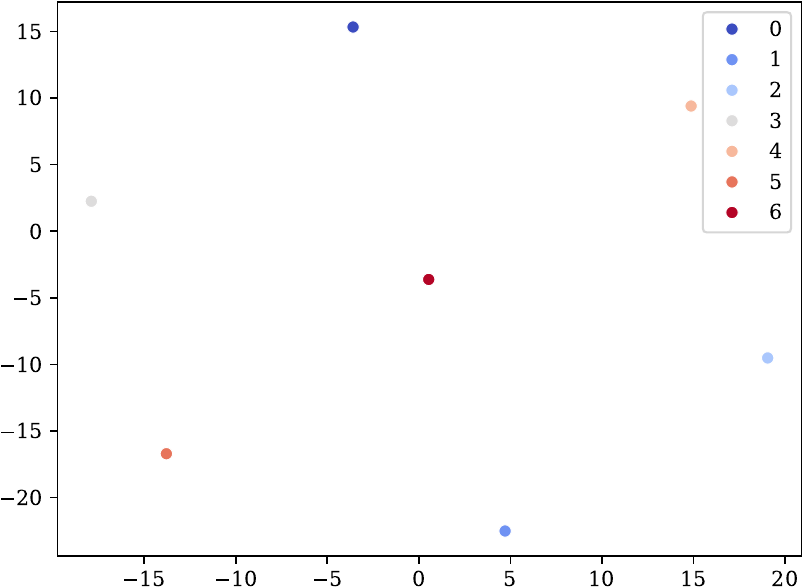}
        \caption{$E_{dow}$.}
        \label{fig:dow}
    \end{subfigure}
    \begin{subfigure}[b]{.3\textwidth}
        \includegraphics[width=\textwidth]{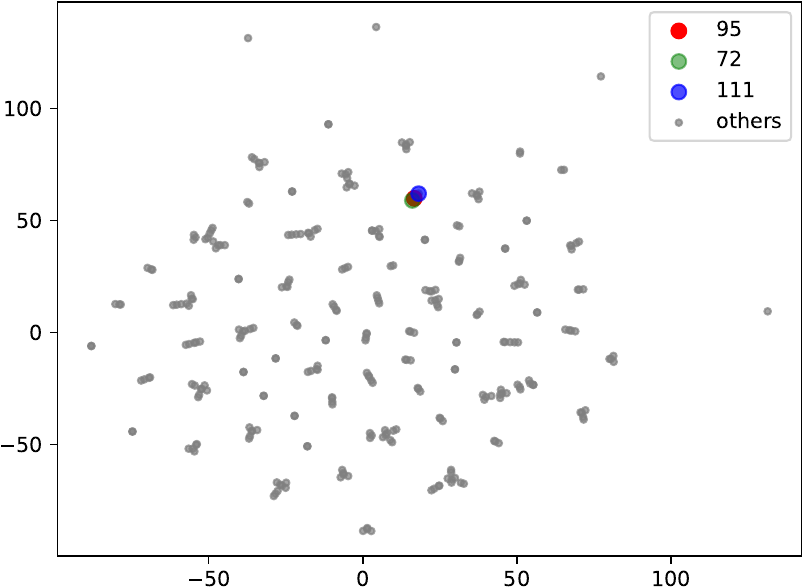}
        \caption{$E_s$.}
        \label{fig:spatialEmb}
    \end{subfigure}
    \caption{Visualizations of $E_{tod}$, $E_{dow}$, and $E_s$.}
    \label{fig:visualizationOfEmbeddings}
\end{figure*}

Extracting the parameter weight of the PEMS04-24-step model checkpoint, we visualize the additional embeddings by reducing them to two dimensions using t-SNE~\cite{tsne1, tsne2}, as shown in Fig.~\ref{fig:tod},~\ref{fig:dow} and~\ref{fig:spatialEmb}. The effectiveness of the additional embeddings is confirmed through visualization experiments.

For $E_{tod}$ in Fig.~\ref{fig:tod}, the color bar indicates the moment from 0:00 to 24:00 with the spectrum that blends from a dark purple into a vivid yellow. The positions of dark purple and bright yellow points are close, indicating that the temporal features in the late night and early morning are similar. More interestingly, the moments represented by light green and dark green points exhibit a symmetrical pattern, fully reflecting the unimodal symmetric trend of the traffic flow rising and falling, as shown in Fig.~\ref{fig:nodeflow}. Overall, points with similar colors are close to each other, but the direction of variation within small regions is irregular. This also aligns with the pattern in Fig.~\ref{fig:nodeflow}, where the overall trend is clear, but small fluctuations are intense. 

To our surprise, the points representing each day in the $E_{dow}$ visualization are almost uniformly distributed around the point for Saturday. We suspect this may be due to insufficient sample size in the dataset (the PEMS04 dataset spans only 8 weeks) to optimize a more effective representation, or it could be that the periodic characteristics of the weekly cycle are not pronounced, with relatively minor variations. The picture aligns with the results of the ablation experiments in Section~\ref{sec:ablation}. However, the ablation study indicates that $E_{dow}$ still contributes somewhat to prediction accuracy, which is why we retain this module.

In the visualization of $E_s$, we highlight nodes 95, 72, and 111 aforementioned in Fig.~\ref{fig:nodeflow} that exhibit similar patterns even though they are not adjacent. The three colorful points almost overlap, indicating that the clustering results reflect the similarity of spatial features rather than merely topological adjacency. This strongly supports our decision to abandon the topology structure as a strong prior knowledge of the encoder. Conversely, if a node focuses exclusively on its neighbors, it loses some of the patterns it could otherwise learn.

\subsection{Error Analysis}\label{sec:erroranalysis}

\begin{figure}[ht]
    \centering
    \includegraphics[width=.9\columnwidth]{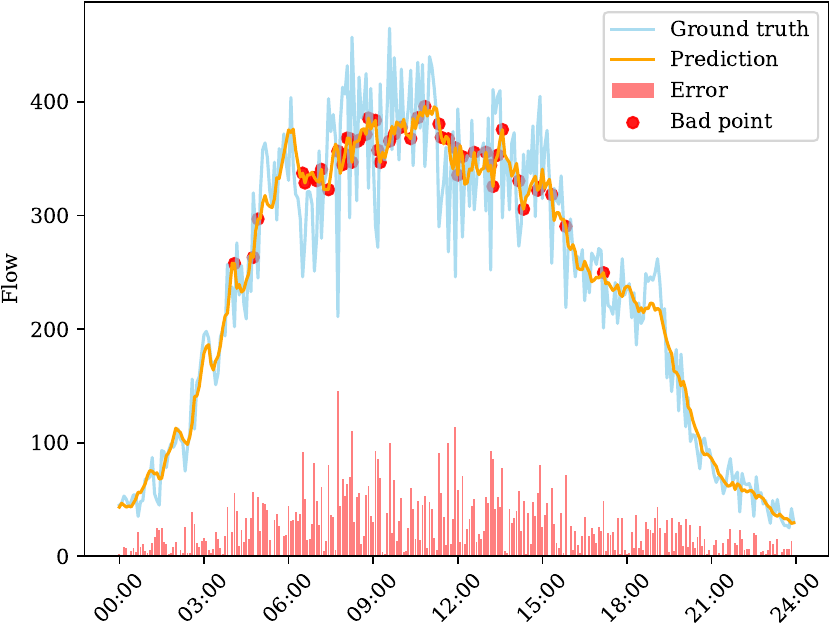}
    \caption{Error analysis. \model~has learned the main trend patterns, but the prediction accuracy reduces during the traffic peak.}
    \label{fig:erroranalysis}
\end{figure}

We conducted an error analysis on the PEMS04 24-step scenario, as shown in Fig.~\ref{fig:erroranalysis}. The threshold for bad points is set at 2.5 times the MAE. The result indicates that \model~has learned the main trend patterns. However, when the ground truth shakes violently, the error rises, which is most remarkable during 9:00 -- 15:00, the peak traffic hours. Honestly, pulse variation is a common phenomenon in traffic flow data in the real world, and it is the key to enhancing performance in future work.

\subsection{Hyperparameter Configuration}\label{sec:hyperparameter}
We report the key hyperparameters in TABLE~\ref{tab:Hyperparameters}, where the mask ratio and subgraph size are selected based on the best-performing settings from the previous ablation study. The embedding dimension is equal across all four embeddings, which are concatenated as input to the encoder, hence the hidden dimension is 4 times the embedding dimension. A linear learning rate decay is used, with the milestone indicating the decay epoch.

\begin{table}[ht]
\centering
\begin{tabular}{l|ccc}
\toprule
\textbf{Hyperparameter} & \textbf{PEMS04} & \textbf{PEMS08} & \textbf{SEATTLE} \\
\midrule
Mask ratio           & 0.2         & 0.2         & 0.2 \\
Subgraph size        & 50          & 30          & 50 \\
Embedding dim        & 64          & 32          & 64 \\
Feed-forward dim     & 1024        & 1024        & 1024 \\
Heads                & 4           & 4           & 4 \\
Layers               & 1           & 1           & 1 \\
Learning rate        & 1e-4        & 1e-4        & 1e-4 \\
LR milestones        & [55]        & [65]        & [55] \\
LR decay rate        & 0.1         & 0.1         & 0.1 \\
Early stop patience  & 10          & 10          & 10 \\
\bottomrule
\end{tabular}

\caption{Key Hyperparameter Settings.}
\label{tab:Hyperparameters}

\end{table}

\section{Limitations and Future Work}\label{sec:limitation}

Our model is fundamentally data-driven, relying solely on patterns learned from observed historical data. While this approach has proven effective at capturing recurring trends and dependencies, it also introduces a critical limitation: the inability to respond to emergent or unprecedented traffic. For example, sudden events such as accidents, road closures, or large-scale public gatherings can significantly alter traffic patterns, deviating from historical norms. This limitation diminishes the practical significance of our predictions for decision-making purposes, particularly in situations where rapid response and adaptability are crucial.

Some spatial-temporal datasets already include more metadata information~\cite{largestst, dataset1terra}. Integrating them into the backbone may involve multi-modal processing. It is believed that incorporating these richer sources of information can provide greater robustness and stability.

TFG embeds a temporal window into each node token, enabling VisiFold to process temporal information solely through the embedding layer. Although this design significantly simplifies the overall architecture without a dedicated temporal module, it is a promising future direction to integrate temporal modules into the VisiFold framework.

Moreover, building upon the success of the new tokenization, we aim to explore its applicability to a broader range of spatial-temporal tasks. This would demonstrate the generalization and efficiency of our proposed method, potentially establishing it as a versatile mechanism for various spatial-temporal tasks.

\section{Conclusion}
VisiFold addresses the core challenges of long-term traffic forecasting: complex spatial-temporal dependencies and high resource cost. Therein, the temporal folding graph folds a sequence of snapshots into a single graph, where temporal dynamics are modeled within nodes, while spatial dependencies are exchanged across nodes. To further improve accuracy and efficiency, we introduce node visibility, which incorporates node-level masking and subgraph sampling. VisiFold alleviates bottlenecks in temporal and spatial dimensions and surpasses strong baselines in both accuracy and efficiency. Remarkably, inference completes in under one second, favoring real-time prediction and edge deployment. Performance remains stable and even exceeds full-graph training when up to 80\% of nodes are masked. Overall, VisiFold is an efficient traffic forecasting framework that we expect to enable longer, more precise, and larger-scale forecasting and to support downstream traffic decision-making.

\section*{Acknowledgments}
This work was supported by Fundamental Research Funds for the Central Universities under Grant 2025JBZX058, National Nature Science Foundation of China (No. 62406020) and Beijing Natural Science Foundation (No. L252034). We thank the reviewers and chairs for their constructive feedback.

\section*{AI-Generated Content Acknowledgment}
Portions of the language and phrasing in this paper were refined with the assistance of a large language model (LLM) to improve clarity and readability. All content, including ideas, experimental results, and conclusions, was conceived and written by the authors.

\bibliographystyle{IEEEtranN}
\bibliography{bibliography}

@article{agcrn,
  title   = {Adaptive graph convolutional recurrent network for traffic forecasting},
  author  = {Bai, Lei and Yao, Lina and Li, Can and Wang, Xianzhi and Wang, Can},
  journal = {Advances in neural information processing systems},
  volume  = {33},
  pages   = {17804--17815},
  year    = {2020}
}

@article{astgnn,
  author   = {Guo, Shengnan and Lin, Youfang and Wan, Huaiyu and Li, Xiucheng and Cong, Gao},
  journal  = {IEEE Transactions on Knowledge and Data Engineering},
  title    = {Learning Dynamics and Heterogeneity of Spatial-Temporal Graph Data for Traffic Forecasting},
  year     = {2022},
  volume   = {34},
  number   = {11},
  pages    = {5415-5428},
  doi      = {10.1109/TKDE.2021.3056502}
}

@article{dcrnn,
  title   = {Diffusion convolutional recurrent neural network: Data-driven traffic forecasting},
  author  = {Li, Yaguang and Yu, Rose and Shahabi, Cyrus and Liu, Yan},
  journal = {arXiv preprint arXiv:1707.01926},
  year    = {2017}
}

@inproceedings{gman,
  author    = {Chuanpan Zheng and Xiaoliang Fan and Cheng Wang and Jianzhong Qi},
  title     = {GMAN: A Graph Multi-Attention Network for Traffic Prediction},
  booktitle = {AAAI},
  pages     = {1234--1241},
  year      = {2020}
}

@inproceedings{graphwarvenet,
  author    = {Wu, Zonghan and Pan, Shirui and Long, Guodong and Jiang, Jing and Zhang, Chengqi},
  title     = {Graph wavenet for deep spatial-temporal graph modeling},
  year      = {2019},
  isbn      = {9780999241141},
  publisher = {AAAI Press},
  booktitle = {Proceedings of the 28th International Joint Conference on Artificial Intelligence},
  pages     = {1907–1913},
  numpages  = {7},
  location  = {Macao, China},
  series    = {IJCAI'19}
}

@inproceedings{sstban,
  title        = {Self-supervised spatial-temporal bottleneck attentive network for efficient long-term traffic forecasting},
  author       = {Guo, Shengnan and Lin, Youfang and Gong, Letian and Wang, Chenyu and Zhou, Zeyu and Shen, Zekai and Huang, Yiheng and Wan, Huaiyu},
  booktitle    = {2023 IEEE 39th International Conference on Data Engineering (ICDE)},
  pages        = {1585--1596},
  year         = {2023},
  organization = {IEEE}
}

@inproceedings{stgcn,
  title     = {Spatio-temporal Graph Convolutional Networks: A Deep Learning Framework for Traffic Forecasting},
  author    = {Yu, Bing and Yin, Haoteng and Zhu, Zhanxing},
  booktitle = {Proceedings of the 27th International Joint Conference on Artificial Intelligence (IJCAI)},
  year      = {2018}
}

@article{STGNNs,
  title     = {A comprehensive survey on graph neural networks},
  author    = {Wu, Zonghan and Pan, Shirui and Chen, Fengwen and Long, Guodong and Zhang, Chengqi and Philip, S Yu},
  journal   = {IEEE transactions on neural networks and learning systems},
  volume    = {32},
  number    = {1},
  pages     = {4--24},
  year      = {2020},
  publisher = {IEEE}
}

@inproceedings{stsgcn,
  title     = {Spatial-temporal synchronous graph convolutional networks: A new framework for spatial-temporal network data forecasting},
  author    = {Song, Chao and Lin, Youfang and Guo, Shengnan and Wan, Huaiyu},
  booktitle = {Proceedings of the AAAI conference on artificial intelligence},
  pages     = {914--921},
  year      = {2020}
}

@inproceedings{pdformer,
  title={PDFormer: Propagation Delay-aware Dynamic Long-range Transformer for Traffic Flow Prediction},
  author={Jiawei Jiang and 
  		  Chengkai Han and 
  		  Wayne Xin Zhao and 
  		  Jingyuan Wang},
  booktitle = {{AAAI}},
  publisher = {{AAAI} Press},
  year      = {2023}
}

@article{transformer,
  title   = {Attention is all you need},
  author  = {Vaswani, Ashish and Shazeer, Noam and Parmar, Niki and Uszkoreit, Jakob and Jones, Llion and Gomez, Aidan N and Kaiser, {\L}ukasz and Polosukhin, Illia},
  journal = {Advances in neural information processing systems},
  volume  = {30},
  year    = {2017}
}

@article{ts-at,
  title   = {Towards Spatio-Temporal Aware Traffic Time Series Forecasting--Full Version},
  author  = {Cirstea, Razvan-Gabriel and Yang, Bin and Guo, Chenjuan and Kieu, Tung and Pan, Shirui},
  journal = {arXiv preprint arXiv:2203.15737},
  year    = {2022}
}

@article{st-transformer,
  title={Spatial-temporal transformer networks for traffic flow forecasting},
  author={Xu, Mingxing and Dai, Wenrui and Liu, Chunmiao and Gao, Xing and Lin, Weiyao and Qi, Guo-Jun and Xiong, Hongkai},
  journal={arXiv preprint arXiv:2001.02908},
  year={2020}
}

@inproceedings{staeformer,
  title={Spatio-temporal adaptive embedding makes vanilla transformer sota for traffic forecasting},
  author={Liu, Hangchen and Dong, Zheng and Jiang, Renhe and Deng, Jiewen and Deng, Jinliang and Chen, Quanjun and Song, Xuan},
  booktitle={Proceedings of the 32nd ACM International Conference on Information and Knowledge Management},
  pages={4125--4129},
  year={2023}
}

@inproceedings{
gcn,
title={Semi-Supervised Classification with Graph Convolutional Networks},
author={Thomas N. Kipf and Max Welling},
booktitle={International Conference on Learning Representations},
year={2017},
url={https://openreview.net/forum?id=SJU4ayYgl}
}

@article{arima,
  title={Modeling and Forecasting Vehicular Traffic Flow as a Seasonal ARIMA Process: Theoretical Basis and Empirical Results},
  author={Billy M. Williams and Lester A. Hoel},
  journal={Journal of Transportation Engineering},
  year={2003},
  volume={129},
  pages={664-672},
  url={https://api.semanticscholar.org/CorpusID:14712092}
}

@article{var,
  title={Integrating Granger Causality and Vector Auto-Regression for Traffic Prediction of Large-Scale WLANs},
  author={Zheng Lu and Chen Zhou and Jing Wu and Hao Jiang and Songyue Cui},
  journal={KSII Trans. Internet Inf. Syst.},
  year={2016},
  volume={10},
  pages={136-151},
  url={https://api.semanticscholar.org/CorpusID:38453985}
}

@inproceedings{dmstgcn,
author = {Han, Liangzhe and Du, Bowen and Sun, Leilei and Fu, Yanjie and Lv, Yisheng and Xiong, Hui},
title = {Dynamic and Multi-faceted Spatio-temporal Deep Learning for Traffic Speed Forecasting},
year = {2021},
isbn = {9781450383325},
publisher = {Association for Computing Machinery},
address = {New York, NY, USA},
url = {https://doi.org/10.1145/3447548.3467275},
doi = {10.1145/3447548.3467275},
booktitle = {Proceedings of the 27th ACM SIGKDD Conference on Knowledge Discovery \& Data Mining},
pages = {547–555},
numpages = {9},
location = {Virtual Event, Singapore},
series = {KDD '21}
}

@inproceedings{astgcn,
  title={Attention based spatial-temporal graph convolutional networks for traffic flow forecasting},
  author={Guo, Shengnan and Lin, Youfang and Feng, Ning and Song, Chao and Wan, Huaiyu},
  booktitle={Proceedings of the AAAI Conference on Artificial Intelligence},
  volume={33},
  pages={922--929},
  year={2019}
}

@inproceedings{autost,
 author = {Li, Jianxin and Zhang, Shuai and Xiong, Hui and Zhou, Haoyi},
 booktitle = {Advances in Neural Information Processing Systems},
 editor = {S. Koyejo and S. Mohamed and A. Agarwal and D. Belgrave and K. Cho and A. Oh},
 pages = {20498--20510},
 publisher = {Curran Associates, Inc.},
 title = {AutoST: Towards the Universal Modeling of Spatio-temporal Sequences},
 volume = {35},
 year = {2022}
}

@article{datasetCollection,
  title={Freeway performance measurement system: mining loop detector data},
  author={Chen, Chao and Petty, Karl and Skabardonis, Alexander and Varaiya, Pravin and Jia, Zhanfeng},
  journal={Transportation Research Record},
  volume={1748},
  number={1},
  pages={96--102},
  year={2001},
  publisher={SAGE Publications Sage CA: Los Angeles, CA}
}

@inproceedings{mae,
  title={Masked autoencoders are scalable vision learners},
  author={He, Kaiming and Chen, Xinlei and Xie, Saining and Li, Yanghao and Doll{\'a}r, Piotr and Girshick, Ross},
  booktitle={Proceedings of the IEEE/CVF conference on computer vision and pattern recognition},
  pages={16000--16009},
  year={2022}
}

@article{huber,
author = {Peter J. Huber},
title = {{Robust Estimation of a Location Parameter}},
volume = {35},
journal = {The Annals of Mathematical Statistics},
number = {1},
publisher = {Institute of Mathematical Statistics},
pages = {73 -- 101},
year = {1964},
doi = {10.1214/aoms/1177703732},
}

@article{adam,
  title={Adam: A method for stochastic optimization},
  author={Kingma, Diederik P and Ba, Jimmy},
  journal={arXiv preprint arXiv:1412.6980},
  year={2014}
}

@inproceedings{geoman,
  title={Geoman: Multi-level attention networks for geo-sensory time series prediction.},
  author={Liang, Yuxuan and Ke, Songyu and Zhang, Junbo and Yi, Xiuwen and Zheng, Yu},
  booktitle={Proceedings of International Joint Conference on Artificial Intelligence},
  pages={3428--3434},
  year={2018}
}

@inproceedings{stmetanet,
  title={Urban traffic prediction from spatio-temporal data using deep meta learning},
  author={Pan, Zheyi and Liang, Yuxuan and Wang, Weifeng and Yu, Yong and Zheng, Yu and Zhang, Junbo},
  booktitle={Proceedings of the 25th ACM SIGKDD Conference on Knowledge Discovery and Data Mining},
  pages={1720--1730},
  year={2019}
}

@inproceedings{stgode,
  title={Spatial-temporal graph ode networks for traffic flow forecasting},
  author={Fang, Zheng and Long, Qingqing and Song, Guojie and Xie, Kunqing},
  booktitle={Proceedings of the 27th ACM SIGKDD Conference on Knowledge Discovery and Data Mining},
  pages={364--373},
  year={2021}
}

@inproceedings{STGNCDE,
  title={Graph neural controlled differential equations for traffic forecasting},
  author={Choi, Jeongwhan and Choi, Hwangyong and Hwang, Jeehyun and Park, Noseong},
  booktitle={Proceedings of the AAAI Conference on Artificial Intelligence},
  pages={6367--6374},
  year={2022}
}

@article{dgcrn,
  title={Dynamic graph convolutional recurrent network for traffic prediction: Benchmark and solution},
  author={Li, Fuxian and Feng, Jie and Yan, Huan and Jin, Guangyin and Yang, Fan and Sun, Funing and Jin, Depeng and Li, Yong},
  journal={ACM Transactions on Knowledge Discovery from Data},
  pages={1--21},
  year={2021},
}

@inproceedings{dstagnn,
  title={Dstagnn: Dynamic spatial-temporal aware graph neural network for traffic flow forecasting},
  author={Lan, Shiyong and Ma, Yitong and Huang, Weikang and Wang, Wenwu and Yang, Hongyu and Li, Pyang},
  booktitle={International Conference on Machine Learning},
  pages={11906--11917},
  year={2022},
}

@inproceedings{ddSTGNN,
  title={Decoupled dynamic spatial-temporal graph neural network for traffic forecasting},
  author={Shao, Zezhi and Zhang, Zhao and Wei, Wei and Wang, Fei and Xu, Yongjun and Cao, Xin and Jensen, Christian S},
  booktitle={Proceedings of the VLDB Endowment},
  pages={2733--2746},
  year={2022}
}

@misc{seattle,
      title={Deep Bidirectional and Unidirectional LSTM Recurrent Neural Network for Network-wide Traffic Speed Prediction}, 
      author={Zhiyong Cui and Ruimin Ke and Ziyuan Pu and Yinhai Wang},
      year={2019},
      eprint={1801.02143},
      archivePrefix={arXiv},
      primaryClass={cs.LG}
}

@misc{character-level,
  title={Character-aware neural language models. CoRR abs/1508.06615},
  author={Kim, Yoon and Jernite, Yacine and Sontag, David and Rush, Alexander M},
  year={2015}
}

@article{word2vec,
  title={Efficient estimation of word representations in vector space},
  author={Mikolov, Tomas and Chen, Kai and Corrado, Greg and Dean, Jeffrey},
  journal={arXiv preprint arXiv:1301.3781},
  year={2013}
}

@article{bpe,
  title={Neural machine translation of rare words with subword units},
  author={Sennrich, Rico and Haddow, Barry and Birch, Alexandra},
  journal={arXiv preprint arXiv:1508.07909},
  year={2015}
}

@article{bert,
  title={Bert: Pre-training of deep bidirectional transformers for language understanding},
  author={Devlin, Jacob},
  journal={arXiv preprint arXiv:1810.04805},
  year={2018}
}

@article{sentencepiece,
  title={Sentencepiece: A simple and language independent subword tokenizer and detokenizer for neural text processing},
  author={Kudo, Taku and Richardson, John},
  journal={arXiv preprint arXiv:1808.06226},
  year={2018}
}

@article{vit,
  title={An image is worth 16x16 words: Transformers for image recognition at scale},
  author={Dosovitskiy, Alexey and Beyer, Lucas and Kolesnikov, Alexander and Weissenborn, Dirk and Zhai, Xiaohua and Unterthiner, Thomas and Dehghani, Mostafa and Minderer, Matthias and Heigold, Georg and Gelly, Sylvain and others},
  journal={arXiv preprint arXiv:2010.11929},
  year={2020}
}

@inproceedings{stid,
author = {Shao, Zezhi and Zhang, Zhao and Wang, Fei and Wei, Wei and Xu, Yongjun},
title = {Spatial-Temporal Identity: A Simple yet Effective Baseline for Multivariate Time Series Forecasting},
year = {2022},
booktitle = {Proceedings of the 31st ACM International Conference on Information \& Knowledge Management},
pages = {4454–4458},
location = {Atlanta, GA, USA}
}

@article{gru,
  title={Empirical evaluation of gated recurrent neural networks on sequence modeling},
  author={Chung, Junyoung and Gulcehre, Caglar and Cho, KyungHyun and Bengio, Yoshua},
  journal={arXiv preprint arXiv:1412.3555},
  year={2014}
}

@article{lstm,
  title={Long short-term memory},
  author={Hochreiter, Sepp and Schmidhuber, J{\"u}rgen},
  journal={Neural computation},
  volume={9},
  number={8},
  pages={1735--1780},
  year={1997},
  publisher={MIT press}
}

@article{hutformer,
  title={Hutformer: Hierarchical u-net transformer for long-term traffic forecasting},
  author={Shao, Zezhi and Wang, Fei and Zhang, Zhao and Fang, Yuchen and Jin, Guangyin and Xu, Yongjun},
  journal={arXiv preprint arXiv:2307.14596},
  year={2023}
}

@inproceedings{stnorm,
  title={ST-Norm: Spatial and Temporal Normalization for Multi-variate Time Series Forecasting},
  author={Deng, Jinliang and Chen, Xiusi and Jiang, Renhe and Song, Xuan and Tsang, Ivor W},
  booktitle={Proceedings of the 27th ACM SIGKDD Conference on Knowledge Discovery \& Data Mining},
  pages={269--278},
  year={2021}
}

@article{stpgnn, title={Spatio-Temporal Pivotal Graph Neural Networks for Traffic Flow Forecasting}, volume={38}, url={https://ojs.aaai.org/index.php/AAAI/article/view/28707}, DOI={10.1609/aaai.v38i8.28707}, number={8}, journal={Proceedings of the AAAI Conference on Artificial Intelligence}, author={Kong, Weiyang and Guo, Ziyu and Liu, Yubao}, year={2024}, month={Mar.}, pages={8627-8635} }

@article{dataset1terra,
  title={Terra: A Multimodal Spatio-Temporal Dataset Spanning the Earth},
  author={Chen, Wei and Hao, Xixuan and Wu, Yuankai and Liang, Yuxuan},
  journal={Advances in Neural Information Processing Systems},
  volume={37},
  pages={66329--66356},
  year={2025}
}

@article{largestst,
  title={Largest: A benchmark dataset for large-scale traffic forecasting},
  author={Liu, Xu and Xia, Yutong and Liang, Yuxuan and Hu, Junfeng and Wang, Yiwei and Bai, Lei and Huang, Chao and Liu, Zhenguang and Hooi, Bryan and Zimmermann, Roger},
  journal={Advances in Neural Information Processing Systems},
  volume={36},
  pages={75354--75371},
  year={2023}
}

@article{tsne1,
  title={Stochastic neighbor embedding},
  author={Hinton, Geoffrey E and Roweis, Sam},
  journal={Advances in neural information processing systems},
  volume={15},
  year={2002}
}

@article{tsne2,
  title={Visualizing data using t-SNE.},
  author={Van der Maaten, Laurens and Hinton, Geoffrey},
  journal={Journal of machine learning research},
  volume={9},
  number={11},
  year={2008}
}

@misc{gelu,
      title={Gaussian Error Linear Units (GELUs)}, 
      author={Dan Hendrycks and Kevin Gimpel},
      year={2023},
      eprint={1606.08415},
      archivePrefix={arXiv},
      primaryClass={cs.LG},
      url={https://arxiv.org/abs/1606.08415}, 
}

@inproceedings{unet,
  title={U-net: Convolutional networks for biomedical image segmentation},
  author={Ronneberger, Olaf and Fischer, Philipp and Brox, Thomas},
  booktitle={Medical image computing and computer-assisted intervention--MICCAI 2015: 18th international conference, Munich, Germany, October 5-9, 2015, proceedings, part III 18},
  pages={234--241},
  year={2015},
  organization={Springer}
}

@InProceedings{resnet,
author = {He, Kaiming and Zhang, Xiangyu and Ren, Shaoqing and Sun, Jian},
title = {Deep Residual Learning for Image Recognition},
booktitle = {Proceedings of the IEEE Conference on Computer Vision and Pattern Recognition (CVPR)},
month = {June},
year = {2016}
}

@inproceedings{patchtst,
  title     = {A Time Series is Worth 64 Words: Long-term Forecasting with Transformers},
  author    = {Nie, Yuqi and
               H. Nguyen, Nam and
               Sinthong, Phanwadee and 
               Kalagnanam, Jayant},
  booktitle = {International Conference on Learning Representations},
  year      = {2023}
}

@article{
totem,
title={{TOTEM}: {TO}kenized Time Series {EM}beddings for General Time Series Analysis},
author={Sabera J Talukder and Yisong Yue and Georgia Gkioxari},
journal={Transactions on Machine Learning Research},
issn={2835-8856},
year={2024},
url={https://openreview.net/forum?id=QlTLkH6xRC},
note={}
}

@article{vqvae,
  title={Neural discrete representation learning},
  author={Van Den Oord, Aaron and Vinyals, Oriol and others},
  journal={Advances in neural information processing systems},
  volume={30},
  year={2017}
}

@article{timexer,
  title={Timexer: Empowering transformers for time series forecasting with exogenous variables},
  author={Wang, Yuxuan and Wu, Haixu and Dong, Jiaxiang and Qin, Guo and Zhang, Haoran and Liu, Yong and Qiu, Yunzhong and Wang, Jianmin and Long, Mingsheng},
  journal={arXiv preprint arXiv:2402.19072},
  year={2024}
}

@inproceedings{crossformer,
  title={Crossformer: Transformer utilizing cross-dimension dependency for multivariate time series forecasting},
  author={Zhang, Yunhao and Yan, Junchi},
  booktitle={The eleventh international conference on learning representations},
  year={2023}
}

@article{longterm1,
  title={Long-term traffic prediction based on lstm encoder-decoder architecture},
  author={Wang, Zhumei and Su, Xing and Ding, Zhiming},
  journal={IEEE Transactions on Intelligent Transportation Systems},
  volume={22},
  number={10},
  pages={6561--6571},
  year={2020},
  publisher={IEEE}
}

@article{longterm2,
  title={Deep architecture for traffic flow prediction: Deep belief networks with multitask learning},
  author={Huang, Wenhao and Song, Guojie and Hong, Haikun and Xie, Kunqing},
  journal={IEEE Transactions on Intelligent Transportation Systems},
  volume={15},
  number={5},
  pages={2191--2201},
  year={2014},
  publisher={IEEE}
}

@article{longterm3,
  title={Repeatability and similarity of freeway traffic flow and long-term prediction under big data},
  author={Hou, Zhongsheng and Li, Xingyi},
  journal={IEEE Transactions on Intelligent Transportation Systems},
  volume={17},
  number={6},
  pages={1786--1796},
  year={2016},
  publisher={IEEE}
}

@article{long4,
  title={Long-term traffic speed prediction based on multiscale spatio-temporal feature learning network},
  author={Zang, Di and Ling, Jiawei and Wei, Zhihua and Tang, Keshuang and Cheng, Jiujun},
  journal={IEEE Transactions on Intelligent Transportation Systems},
  volume={20},
  number={10},
  pages={3700--3709},
  year={2018},
  publisher={IEEE}
}

@inproceedings{long5,
  title={STCNN: A spatio-temporal convolutional neural network for long-term traffic prediction},
  author={He, Zhixiang and Chow, Chi-Yin and Zhang, Jia-Dong},
  booktitle={2019 20th IEEE international conference on mobile data management (MDM)},
  pages={226--233},
  year={2019},
  organization={IEEE}
}

@article{long6,
title = {Long-term traffic flow forecasting using a hybrid CNN-BiLSTM model},
journal = {Engineering Applications of Artificial Intelligence},
volume = {121},
pages = {106041},
year = {2023},
issn = {0952-1976},
doi = {https://doi.org/10.1016/j.engappai.2023.106041},
url = {https://www.sciencedirect.com/science/article/pii/S0952197623002257},
author = {Manuel Méndez and Mercedes G. Merayo and Manuel Núñez},
}

@inproceedings{stdn,
  title={Spatiotemporal-aware Trend-Seasonality Decomposition Network for Traffic Flow Forecasting},
  author={Cao, Lingxiao and Wang, Bin and Jiang, Guiyuan and Yu, Yanwei and Dong, Junyu},
  booktitle={Proceedings of the AAAI Conference on Artificial Intelligence},
  volume={39},
  number={11},
  pages={11463--11471},
  year={2025}
}

@INBOOK{rnn,
  author={Rumelhart, David E. and McClelland, James L.},
  booktitle={Parallel Distributed Processing: Explorations in the Microstructure of Cognition: Foundations}, 
  title={Learning Internal Representations by Error Propagation}, 
  year={1987},
  volume={},
  number={},
  pages={318-362},
  doi={}}

@article{dynamicgraphSurvey,
  title={Dynamic graph representation learning with neural networks: A survey},
  author={Yang, Leshanshui and Chatelain, Clement and Adam, Sebastien},
  journal={Ieee Access},
  volume={12},
  pages={43460--43484},
  year={2024},
  publisher={IEEE}
}

@inproceedings{graph_joco,
  title={Graph contrastive learning automated},
  author={You, Yuning and Chen, Tianlong and Shen, Yang and Wang, Zhangyang},
  booktitle={International conference on machine learning},
  pages={12121--12132},
  year={2021},
  organization={PMLR}
}

@article{graph-2,
  title={Graph contrastive learning with augmentations},
  author={You, Yuning and Chen, Tianlong and Sui, Yongduo and Chen, Ting and Wang, Zhangyang and Shen, Yang},
  journal={Advances in neural information processing systems},
  volume={33},
  pages={5812--5823},
  year={2020}
}

@inproceedings{qiu2020gcc,
  title={Gcc: Graph contrastive coding for graph neural network pre-training},
  author={Qiu, Jiezhong and Chen, Qibin and Dong, Yuxiao and Zhang, Jing and Yang, Hongxia and Ding, Ming and Wang, Kuansan and Tang, Jie},
  booktitle={Proceedings of the 26th ACM SIGKDD international conference on knowledge discovery \& data mining},
  pages={1150--1160},
  year={2020}
}

@inproceedings{jin2023trafformer,
  title={Trafformer: Unify time and space in traffic prediction},
  author={Jin, Di and Shi, Jiayi and Wang, Rui and Li, Yawen and Huang, Yuxiao and Yang, Yu-Bin},
  booktitle={Proceedings of the AAAI conference on artificial intelligence},
  volume={37},
  number={7},
  pages={8114--8122},
  year={2023}
}

\end{document}